\documentclass[10pt,twocolumn,letterpaper]{article}
\usepackage{multirow}
\usepackage{dsfont}
\usepackage{epigraph} 
\usepackage{siunitx}
\usepackage[pagenumbers]{iccv}              

\renewcommand{\paragraph}[1]{\vspace{2pt plus 1pt minus 1pt}\noindent{\bf #1}\;}
\definecolor{cvprblue}{rgb}{0.21,0.49,0.74}
\usepackage[pagebackref,breaklinks,colorlinks,allcolors=cvprblue]{hyperref}

\def\confName{CVPR}

\title{TrafficLoc: Localizing  Traffic Surveillance Cameras in 3D Scenes}

\author{
	\small
	\begin{tabular}{c c c c c c}                                            
		Yan Xia$^{*1,2 \footnotemark[2]}$ &
		Yunxiang Lu$^{*1}$ &
		Rui Song$^1$ &
            Oussema Dhaouadi$^{1,3}$ & 
		João F. Henriques$^4$ &
		Daniel Cremers$^{1,2}$ \\                                        
		\multicolumn{6}{c}{\shortstack{$^1$Technical University of Munich $^2$Munich Center for Machine Learning (MCML) \\ $^3$ DeepScenario $^4$ Visual Geometry Group, University of Oxford  }} \\                                                
		\multicolumn{6}{c}{\{yan.xia, yunxiang.lu, rui.song,
        oussema.dhaouadi,
        cremers\}@tum.de, joao@robots.ox.ac.uk } 
	\end{tabular}                                                                       
}

\begin{document}
\twocolumn[{%
\renewcommand\twocolumn[1][]{#1}%
\maketitle
\begin{center}
\vspace{-0.5cm}
\captionsetup{type=figure}
\includegraphics[width=0.99\linewidth]{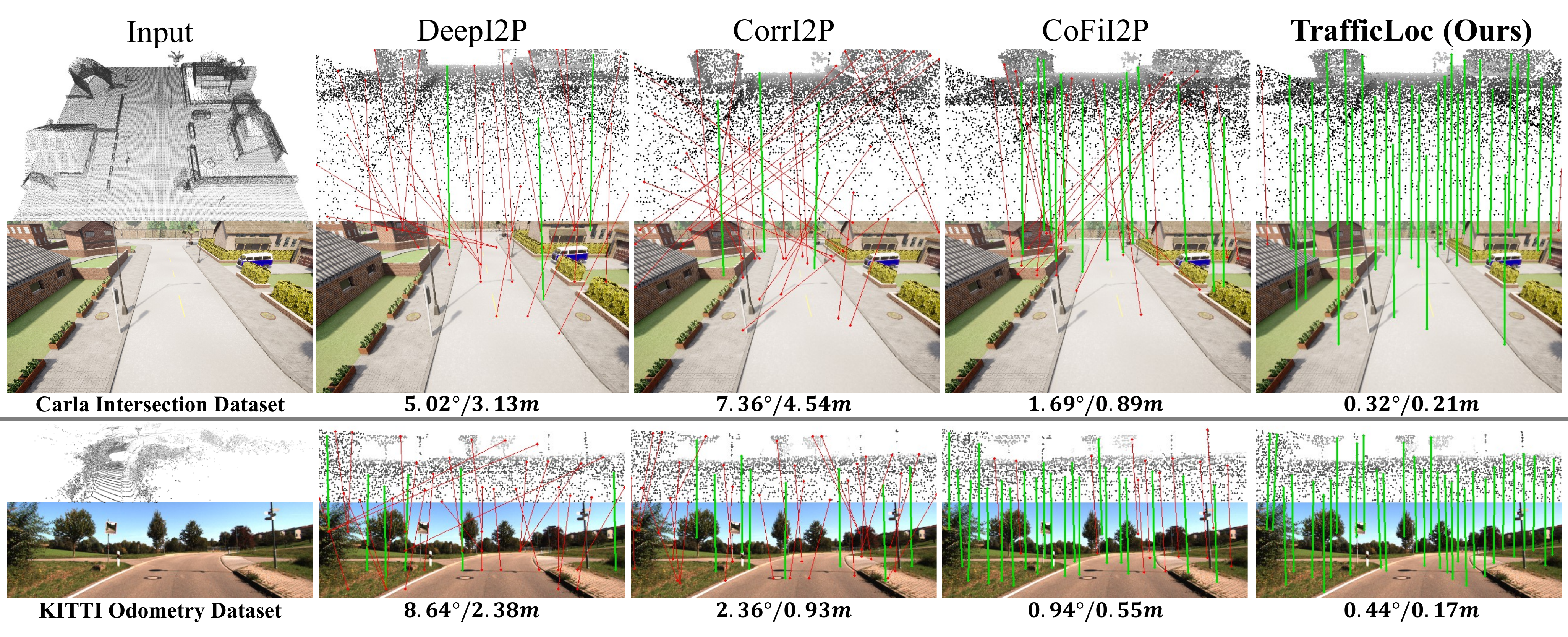}
\captionof{figure}{Localization accuracy on the proposed \textit{Carla Intersection} and \textit{KITTI} dataset. 
The point cloud is projected into a 2D view and shown above the image, with point colors indicating distance. 
The proposed \textit{TrafficLoc}
achieves better performance, with more correct (\textcolor{green}{green}) and fewer incorrect (\textcolor{red}{red}) point-to-pixel pairs. 
The first column presents the input point cloud and input image.
}

\label{fig:teaser}
\end{center}%
}]

\let\oldthefootnote\thefootnote
\renewcommand{\thefootnote}{\fnsymbol{footnote}} 
\footnotetext[2]{Corresponding author. * Equal contribution.} 
\let\thefootnote\oldthefootnote

\begin{abstract}
We tackle the problem of localizing traffic cameras within a 3D reference map and propose a novel image-to-point cloud registration (I2P) method, TrafficLoc, in a coarse-to-fine matching fashion.
To overcome the lack of large-scale real-world intersection datasets, we first introduce Carla Intersection, a new simulated dataset with 75 urban and rural intersections in Carla. 
We find that current I2P methods struggle with cross-modal matching under large viewpoint differences, especially at traffic intersections. TrafficLoc thus employs a novel Geometry-guided Attention Loss (GAL) to focus only on the corresponding geometric regions under different viewpoints during 2D-3D feature fusion. 
To address feature inconsistency in paired image patch-point groups, we further propose Inter-intra Contrastive Learning (ICL) to enhance separating 2D patch / 3D group features within each intra-modality and introduce Dense Training Alignment (DTA) with soft-argmax for improving position regression. 
Extensive experiments show our TrafficLoc greatly improves the performance over the SOTA I2P methods (up to \textbf{86\%}) on Carla Intersection and generalizes well to real-world data.
TrafficLoc also achieves new SOTA performance on KITTI and NuScenes datasets, demonstrating the
superiority across both in-vehicle and traffic cameras.
Our project page is publicly available at \url{https://tum-luk.github.io/projects/trafficloc/}.
\end{abstract}    
\section{Introduction}
\label{sec:intro}

Traffic surveillance cameras, as affordable and easy-to-install roadside sensors in cooperative perception, offer a broad, global perspective on traffic.
Integrating this data with onboard sensors enhances situational awareness, supporting applications like early obstacle detection~\cite{ravi2018object_detection,xia2023lightweight} and vehicle localization~\cite{vuong2024toward}.
Localizing the 6-DoF pose (\ie position and orientation) of each traffic camera within a 3D point cloud (e.g., from LiDARs) captured at different locations is thus essential for cooperative perception.

Recent advances in 3D sensors have enhanced Image-to-Point cloud (I2P) registration methods~\cite{continue_line,koide2023general, ren2022corri2p, kang2023cofii2p, yao2023cfi2p}, as LiDAR-acquired point clouds provide accurate and detailed 3D information~\cite{xia2021soe,xia2023casspr}. 
However, unlike these I2P methods, the registration of the traffic camera focuses on determining \textbf{fixed} camera poses within the point cloud scene.
The main challenge of localizing traffic cameras lies in several aspects: 1) Images and 3D reference point clouds are captured at different times and from largely different viewpoints, making it difficult to obtain the precise initial guess required for traditional registration methods~\cite{sheng2024rendering}. 2) Directly projecting reference point clouds onto the images can lead to a ``bleeding problem''~\cite{koide2023general}. 3) Variable focal length cameras are generally used as traffic cameras for
easy installation, causing the intrinsic parameters to be frequently changed during operation~\cite{jing2022intrinsic}. 

To date, only a few methods have been proposed for localizing traffic cameras in the 3D reference scene.
\cite{jing2022intrinsic} employs manual 2D-3D feature matching followed by optimization using distance transform. \cite{vuong2024toward} uses panoramic images for point cloud reconstruction and then aligns traffic camera images with the resulting point cloud. 
~\cite{sheng2024rendering} proposes to automate traffic image-to-point cloud registration by generating synthesized views from point clouds to reduce modality gaps. 
Although these methods achieve promising results, the requirement for manual intervention or panoramic/rendering image acquisition complicates their deployment. It is therefore important to develop the capability to perform traffic camera localization directly. 

However, most of the existing datasets (\ie KITTI~\cite{kitti} and NuScenes~\cite{NuScenes}) are limited to the in-vehicle cameras, lacking sufficient intersections and traffic cameras. 
To address this gap, we first introduce \textit{Carla Intersection}, a new intersection dataset using the Carla simulator. 
\textit{Carla Intersection} provides 75 intersections across 8 worlds, covering both urban and rural landscapes. 
Furthermore, we propose TrafficLoc, a novel I2P registration method for localizing traffic cameras within a 3D reference map. TrafficLoc follows a coarse-to-fine localization strategy, beginning with image patch-to-point group matching and refining localization through pixel-to-point matching.

We find that simply applying transformers for cross-modal feature fusion in current I2P registration methods~\cite{yao2023cfi2p, kang2023cofii2p} struggle with limited geometric awareness and weak robustness to viewpoint variations. To address these challenges, we propose a novel Geometry-guided Feature Fusion (GFF) module, a Transformer-based architecture optimized by a novel Geometry-guided Attention Loss (GAL).
GAL directs the model to focus on geometric-related regions during feature fusion, significantly improving performance in scenarios with large viewpoint changes. 
Another observation is that, when registering image patch and point group features in the coarse matching, the normal contrastive learning in previous I2P methods ignores one fact: \textit{Paired 2D patch-3D point group relationships require separable features within each intra-modality}.  
To address this issue, we propose a novel Inter-intra Contrastive Learning (ICL) to enhance feature distinctiveness between 2D patches and 3D groups.  
Moreover, we find that only aligning features with sparsely paired 2D patch-3D groups potentially neglects extra global features. We thus propose a Dense Training Alignment (DTA) to back-propagate the calculated gradients to all image patches using a soft-argmax operator. These operations enable one-stage training to estimate accurate pixel-point correspondences.


To summarize, the main contributions of this work are:
\begin{itemize}
    \item We set up a new simulated intersection dataset, \textit{Carla Intersection}, to study the traffic camera localization problem in varying environments. \textit{Carla Intersection} includes 75 intersections across 8 worlds, covering both urban and rural landscapes.
    \item We propose a novel I2P registration method, TrafficLoc, following a coarse-to-fine pipeline. A novel Geometry-guided Attention Loss (GAL) is proposed to direct the model to focus on the geometric-related regions during cross-modality feature fusion.
    \item We propose a novel Inter-intra Contrastive Learning (ICL) and a Dense Training Alignment (DTA) with a soft-argmax operator to achieve more fine-grained features when aligning each image patch-point group.
    \item We conduct extensive experiments on the proposed \textit{Carla Intersection}, USTC intersection~\cite{sheng2024rendering}, KITTI~\cite{kitti} and NuScenes~\cite{NuScenes} datasets, showing the proposed TrafficLoc greatly improves over the state-of-the-art methods and generalizes well to real intersection data and in-vehicle camera localization task.
\end{itemize}

\section{Related Work}
\label{sec:related work}

\noindent


\noindent
\textbf{Uni-modal registration.}
Image-to-image registration (I2I) methods aim to establish pixel-level correspondences between different images, enabling the estimation of relative transformations via classical algorithms such as PnP \cite{epnp}. This serves as a fundamental prerequisite for triangulation in Structure-from-Motion (SfM) \cite{sfm, sfm1, sfm2} and SLAM \cite{orbslam}.
Traditional methods use handcrafted descriptors such as SIFT \cite{sift} or ORB \cite{rublee2011orb} to detect and match keypoints. Recent learning-based methods \cite{superpoint, d2net, r2d2} significantly improve the matching robustness against viewpoint variation. 
SuperGlue \cite{superglue} and LightGlue \cite{lightglue} leverage graph neural networks (GNN) and attention mechanisms to refine feature correspondences. Detector-free method LoFTR \cite{sun2021loftr} adopt a coarse-to-fine manner with Transformer to produce semi-dense matches. Efficient LoFTR \cite{wang2024efficient} introduces a lightweight aggregated attention module for further efficiency improvement.
Conventional point cloud-to-point cloud registration (P2P) methods mainly use ICP \cite{ICP1} and its variants \cite{ICP_variants, ICP2}, which highly rely on initial guess under large relative transformation. Recent learning-based methods ~\cite{deepicp, 3dmatch, huang2021predator, yu2021cofinet, ao2023buffer} extract and match point descriptors, and apply robust estimation methods like RANSAC \cite{pnpsolver} to optimize rigid transformation.
Geotransformer \cite{qin2023geotransformer} proposes a geometric transformer to learn transformation-invariant point descriptors in the superpoint matching module.
EYOC \cite{liu2024extend} trains a feature extractor using progressively distant point cloud pairs and can adapt to new data distributions without pose labels.
While uni-modal registration methods are effective for feature matching within a single modality, they struggle to generalize to cross-modal scenarios due to inherent feature discrepancies.

\noindent
\textbf{Image-to-Point cloud registration.} To estimate the relative pose between an image and a point cloud, methods like 2D3D-MatchNet \cite{2d3dmatchnet} and LCD \cite{lcd} use deep networks to learn descriptors jointly from 2D image patches and 3D point cloud patches.
3DTNet \cite{3dtnet} learns 3D local descriptors by integrating 2D and 3D local patches, treating 2D features as auxiliary information. 
Cattaneo \etal \cite{2d3dmatch1} establish a shared global feature space between 2D images and 3D point clouds using a teacher-student model.
Recent VXP~\cite{li2024vxp} improves the retrieval performance by enforcing local similarities in a self-supervised manner.
DeepI2P~\cite{deepi2p} reformulates cross-matching as a classification task, identifying if a projected point lies within an image frustum. FreeReg \cite{wang2023freereg} employs pretrained diffusion models and monocular depth estimators to unify image and point cloud features, enabling single-modality matching without training. EP2P-Loc \cite{kim2023ep2p} performs 2D patch classification for each 3D point in retrieved sub-maps, using positional encoding to determine precise 2D-3D correspondences. CorrI2P~\cite{ren2022corri2p} directly matches dense per-pixel/per-point features in overlapping areas to establish I2P correspondences, while CoFiI2P~\cite{kang2023cofii2p} and CFI2P~\cite{yao2023cfi2p} employ a coarse-to-fine strategy, integrating high-level correspondences into low-level matching to filter mismatches. Recent VP2P~\cite{zhou2024differentiable} proposes an end-to-end I2P network with a differentiable PnP solver. However, all methods are limited to in-vehicle camera images. In this work, our method generalizes well to car and traffic camera perspectives. 
\section{Problem Statement}
Let $\mathbf{I} \in \mathbb{R}^{H \times W \times 3}$ with a resolution of $H \times W$ be a RGB image captured from the traffic cameras. Let $\mathbf{P} \in \mathbb{R}^{N \times 3}$ be a 3D scene point cloud collected by LiDARs at different locations, where $N$ is the number of points. 
The camera intrinsic matrix $\mathbf{K} \in \mathbb{R}^{3 \times 3}$ is known or also can be estimated by DUSt3R~\cite{wang2024dust3r} for initialization.
Our goal, consistent with previous I2P registration methods~\cite{yao2023cfi2p,kang2023cofii2p}, is to estimate the 6-DoF relative transformation $\mathbf{T}=[\mathbf{R} | \mathbf{t}]$ between the image $\mathbf{I}$ and point cloud $\mathbf{P}$, including rotation matrix $\mathbf{R} \in \mathbf{SO}(3)$ and translation vector $\mathbf{t} \in \mathbb{R}^{3}$. 
To achieve this, we design a neural network to learn a function
$\mathcal{F(\bullet)}$ that predicts the set of accurate point-to-pixel correspondences $\mathbf{C^*}$ 
given $\mathbf{I}$ and $\mathbf{P}$. We then use EPnP-RANSAC~\cite{epnp,pnpsolver} to get $\mathbf{T}$ based on the estimated $\mathbf{C^*}$ and $\mathbf{K}$.

\section{Our TrafficLoc}

\begin{figure*}
    \centering
    \includegraphics[width=0.9\linewidth]{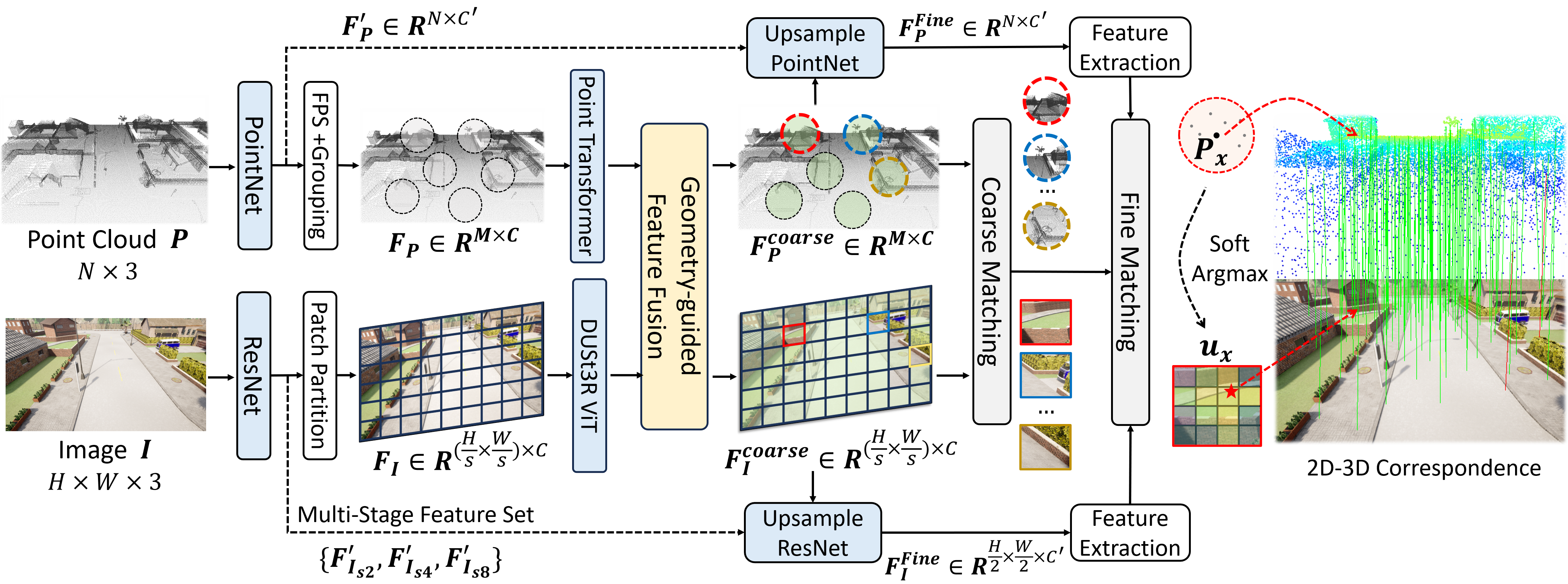}
    \caption{Pipeline of our TrafficLoc. Given a traffic camera image and a 3D scene point cloud collected at different locations, we first extract features at the point group level and image patch level, respectively. We then fuse them using a Geometry-guided Feature Fusion (GFF) module and match them based on similarity rules. Furthermore, we perform fine matching between the point group center and the extracted image window with a soft-argmax operation. Finally,  we use EPnP-RANSAC~\cite{epnp, pnpsolver} algorithm to get the final camera pose based on the predicted 2D-3D correspondences.}
    \label{fig:network}
    \vspace{-0.3cm}
\end{figure*}


Fig. \ref{fig:network} shows our TrafficLoc pipeline.
Given a query image and a reference 3D point cloud, we  
first
extract 2D patch features and 
3D group features respectively, describied in Section~\ref{sec: fe}. Next, we fuse the 2D patch and 3D group feature via a carefully designed GFF module, which will be explained in Section~\ref{sec: geocros}. Section~\ref{sec:cm} describes the coarse matching stage, establishing 2D patch-3D group correspondences. 
Furthermore, we match each 3D group center feature and the
patch features generated from coarse matching in fine matching (Section.~\ref{sec:fm}). Lastly, the EPnP-RANSAC module in Section.~\ref{sec:pe} exploits the point-pixel correspondences set to optimize the relative transformation.

\subsection{Feature Extraction}
\label{sec: fe}
Following~\cite{kim2023ep2p}, we explore a dual branch to extract image and point cloud features using Transformer-based encoders.

\noindent
\textbf{2D patch descriptors.} 
We first utilize ResNet-18~\cite{he2016deep} to extract multi-level features of image $I$ and then use the feature at the coarsest resolution to generate 2D patch descriptors $\mathbf{F_{patch}} \in \mathbb{R}^{HW/s^2 \times C}$, where s is the resolution of non-overlapping patches and $HW/s^2$ is the number of image patches. 
To further enhance the spatial relationships within these image patches, we leverage the pre-trained ViT encoder in DUSt3R ~\cite{wang2024dust3r} owing to its strong capability in regressing 3D coordinates. We reduce the original 1024-dimensional output from DUSt3R to 256 by employing a multi-layer convolutional network with $1\times1$ kernels.

\noindent
\textbf{3D point group descriptors.}
We first utilize a standard PointNet~\cite{pointnet} to extract point-wise features $\mathbf{F^{\prime}_{P}} \in \mathbb{R}^{N \times C^\prime}$ for each point. Then, we conduct Farthest Point Sampling (FPS) to generate $M$ super-points $ \mathbf{Q} = \{\mathbf{q}_1, \mathbf{q}_2, \ldots, \mathbf{q}_{M}\} $ and assign every point in $\mathbf{P}$ to its nearest center in $\mathbf{Q}$, formulating $M$ point groups $G^P = \{ G_1^P, G_2^P, \dots, G_{M}^P \}$ and their associated feature sets $G^{F'} = \{ G_1^{F'}, G_2^{F'}, \dots, G_{M}^{F'} \}$.
Following CFI2P~\cite{yao2023cfi2p}, we then use the Point Transformer~\cite{point_transformer} to enhance local geometry features and incorporate global contextual relationships for each point group.

\subsection{Geometry-guided Feature Fusion}
\label{sec: geocros}
\begin{figure*}
    \centering
    \includegraphics[width=0.9\linewidth]{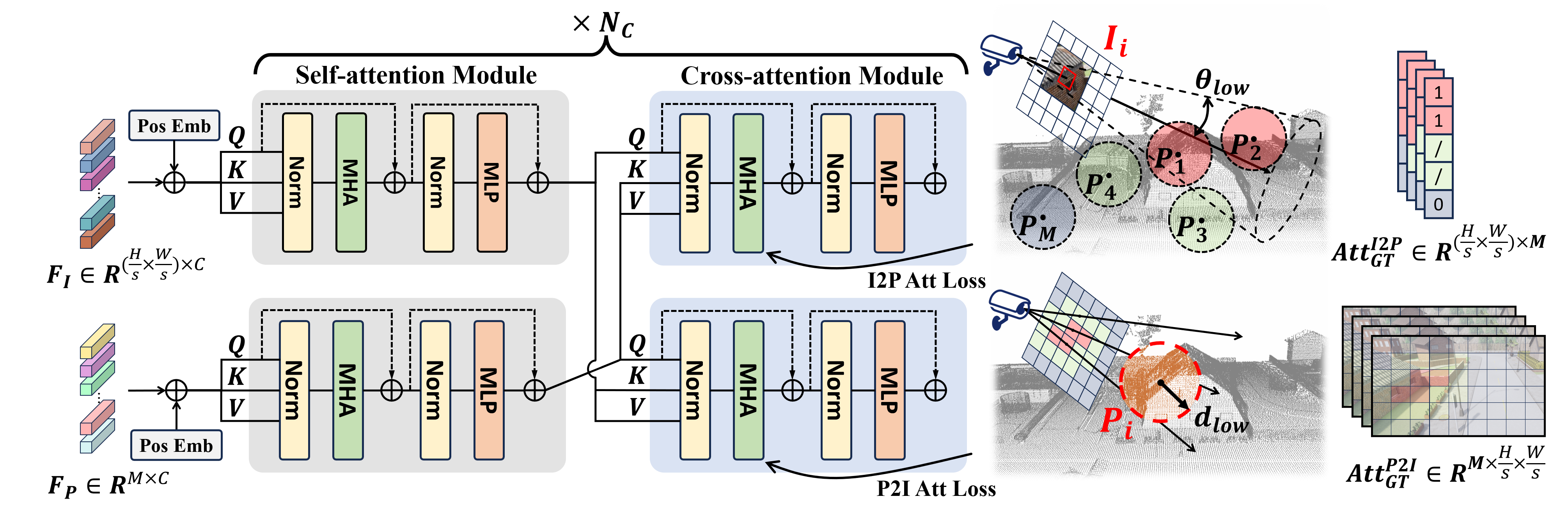}
    \caption{The pipeline of Geometry-guided Feature Fusion (GFF) module. GFF first use $N_c$ layers of self and cross-attention module to enhance the feature across different modalities (left). The proposed Geometry-guided Attention Loss is applied to the cross-attention map of the last fusion layer based on camera projection geometry (right).}
    \label{fig:att}
    \vspace{-0.3cm}
\end{figure*}

Although the I2P registration approaches~\cite{deepi2p, kang2023cofii2p} achieved notable fusion success using a Transformer-based architecture, the inherent differences between different modalities and viewpoints are not well-explored. 
To address this issue, we introduce a novel Geometry-guided Feature Fusion (GFF) module to enhance the network's robustness to viewpoint variations during cross-modal feature fusion. Specifically, we design a Fusion Transformer architecture guided by a novel geometry-guided attention loss (GAL). Fig.~\ref{fig:att} shows the detailed architecture of the GFF module.

\noindent
\textbf{Fusion Transformer.} Similar to ~\cite{yao2023cfi2p}, we adopt a Transformer-based architecture with self-attention and cross-attention layers for cross-modal feature fusion. 
Given the image feature $\mathbf{F}_I$ and point cloud feature $\mathbf{F}_P$, we begin by adding sinusoidal positional embeddings to retain the spatial information within both modalities. In the self-attention module, a transformer encoder enhances features in each modality individually using standard scalar dot-product attention. 
The cross-attention layer is designed to fuse image and point cloud features by applying the attention mechanism across modalities. 
This design allows for the exchange of geometric and textural information between image and point cloud features, enabling a richer, modality-aware feature representation. More details are in \textit{Supp}.

\noindent
\textbf{Geometry-guided Attention Loss (GAL).}
We find that directly applying Transformers for cross-modal feature fusion suffers from limited geometric awareness and weak robustness to viewpoint variations. We thus propose a novel geometry-guided attention loss that supervises cross-modal attention map during training based on geometric alignment, encouraging features to focus on their geometrically corresponding parts, as shown in Fig. \ref{fig:att} (right).

Inspired by~\cite{bhalgat2023light}, we apply supervision to the cross-attention layer at the final stage of the Fusion Transformer, leveraging camera projection geometry as guidance.
In Image-to-Point cloud (I2P) attention, we encourage the network to focus on relevant 3D point groups for each 2D patch feature $I_i$. This is achieved by penalizing attention values that have high values outside the target region and encouraging high attention within the desired area. We implement this through a Binary Cross Entropy (BCE) loss on the \textit{raw} cross-attention map $ATT_{I2P}$:
\begin{equation}
\vspace{-0.1cm}
    \begin{aligned}
    & L_{I2P}(i,j) =\mathrm{BCE}(\sigma(ATT_{I2P}(i,j)),\mathds{1}_{I2P}(i,j)) \\
    & ATT_{I2P} = {F}_I {W}_{Q}\left({F}_P {W}_{K}\right)^{\top} \in {R}^{HW/s^2 \times M}
    \end{aligned}
    \vspace{-0.1cm}
\end{equation}
where $\sigma$ is a sigmoid function, ${W}_{Q}$, ${W}_{K}$ denote the learnable query and key matrices of Transformer, and $\mathds{1}_{I2P}(i,j)$ is a special indicator function defined as:
\vspace{-0.1cm}
\begin{equation}
    \mathds{1}_{I2P}(i, j) = \left\{
    \begin{array}{ll}
        1, & \textrm{if Rad}(i,j) < \theta_{low} \\
        0, & \textrm{if Rad}(i,j) > \theta_{up} \\
    \end{array}
    \right.
\end{equation}
The angular radius Rad$(i,j)$ is given by $\mathrm{Rad}(i,j) = \arccos(\frac{{OI}_i \cdot {OP}_j}{\|{OI}_i\| \|{OP}_j\|})$.
Here, $\mathds{1}_{I2P}(i,j)$ assigns a value of 1 when the angle between the camera ray ${OI_i}$ (formed by the 2D patch feature ${I_i}$ and the camera center ${O}$) and the line to the 3D point center ${P}_j$ is below a threshold $\theta_{low}$. If this angle exceeds an upper threshold $\theta_{up}$, the value is set to 0, indicating that the point is outside the target region. Points with angles between these thresholds are NOT supervised during training, allowing the network to flexibly learn the relationship between attention in these intermediate cases.
Similarly, in Point cloud-to-Image (P2I) attention, we encourage each point group to focus on image patches within its target area of influence:
\begin{equation}
    \begin{aligned}
    L_{P2I}(i,j) & =\mathrm{BCE}(\sigma(ATT_{P2I}(i,j)),\mathds{1}_{P2I}(i,j)) \\
    ATT_{P2I} & = {F}_P {W}_{Q}\left({F}_I {W}_{K}\right)^{\top} \in {R}^{M \times HW/s^2} \\
    \mathds{1}_{P2I}(i,j) & = \left\{
    \begin{array}{ll}
        1 & \textrm{if Dist}(P_i, {O}I_j) < \mathbf{d}_{low} \\
        0 & \textrm{if Dist}(P_i, {O}I_j) > \mathbf{d}_{up} \\
    \end{array}
    \right.
    \end{aligned}
\end{equation}
where Dist$(P_i,{O}I_j)$ represents the distance from point center $P_i$ to the camera ray ${O}I_j$ (formed by the camera center ${O}$ and the patch $I_j$) in the 3D space. Since all point groups share a similar receptive field during feature extraction, the distance threshold is set the same for each. This ensures point groups farther from the camera focus on a smaller target area within the image, aligning with the natural physical principle that closer objects appear larger while distant ones appear smaller.
The final geometry-guided attention loss is:
\begin{equation}
\vspace{-0.1cm}
    L_{Att} = \sum_{i=1}^{HW/s^2} \sum_{j=1}^{M} L_{I2P}(i,j) + L_{P2I}(j,i)
\vspace{-0.1cm}
\end{equation}
Note that the relative GT transformation matrix $T_{GT}$ is required only during training. Experiments in Section~\ref{sec: ablation_study} shows the effectiveness of GAL, leading to significant improvements in scenarios with large viewpoint changes.


\begin{figure}
    \centering
    \includegraphics[width=1.0\linewidth]{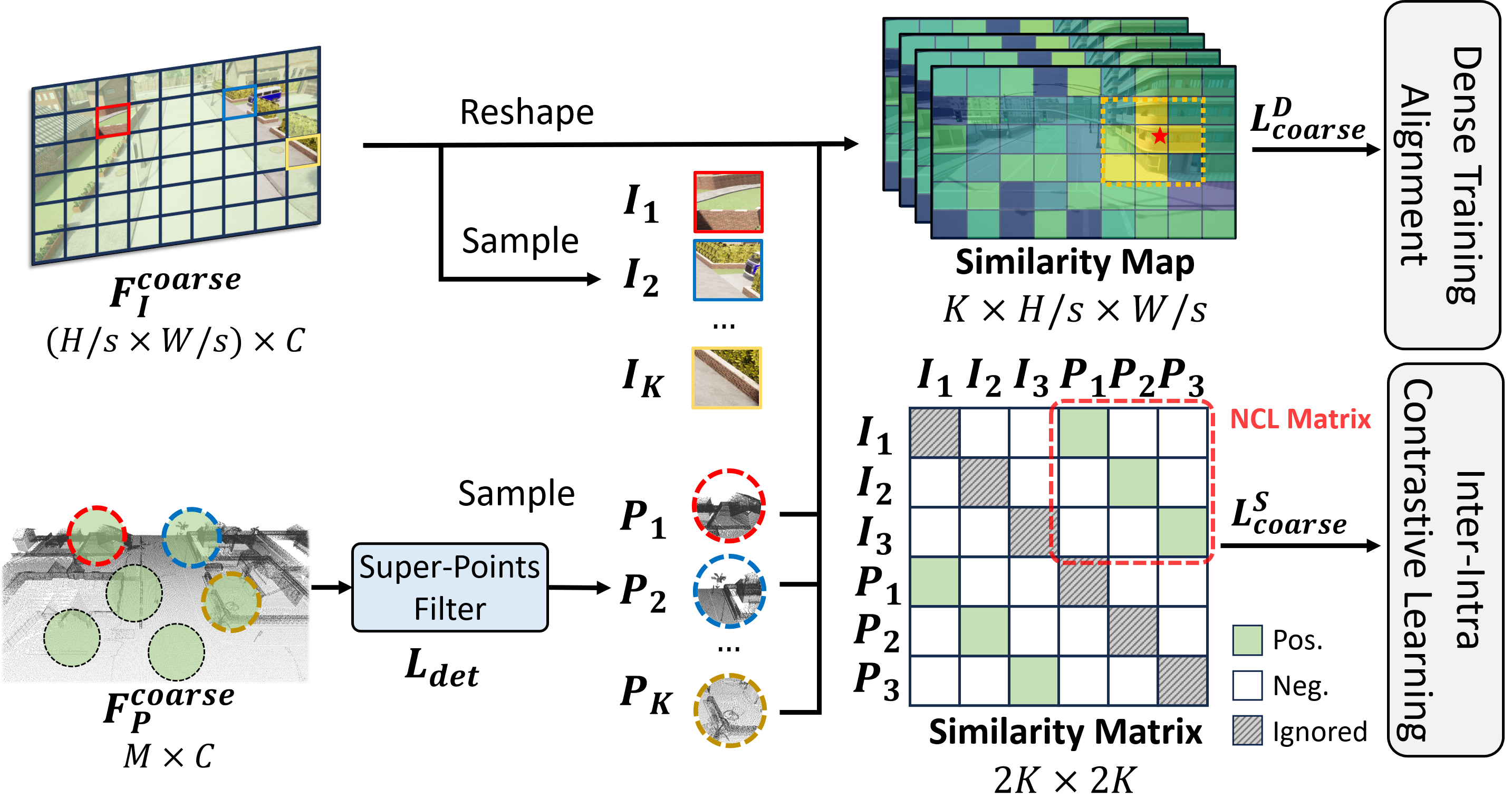}
    \caption{Coarse matching mechanism of TrafficLoc. The positive feature pairs are generated based on ground-truth transformation matrix. The coarse image feature ${F}^{coarse}_I$ is reshaped to compute its similarity map with each coarse point feature.}
    \label{fig:coarse_matching}
    \vspace{-0.5cm}
\end{figure}

\subsection{Coarse Matching}
~\label{sec:cm}
Here we aim to match the point group with the image patch at the coarse level given the fused 2D patch features $\{{F}^{coarse}_I\}^{N_p}_{i=1}$ and 3D point group features $\{{F}^{coarse}_P\}^{N_g}_{i=1}$. 
Since a monocular camera can only capture a part of the 3D point cloud scene due to the limited field of view (FoV), we first apply a simple super-points filter with binary classification MLP head to predict super-points in or beyond the frustum, supervised by In-frustum Detection Loss $L_{det}$. 
For each predicted in-frustum point group $P_i$, we estimate its corresponding coarse pixel $u_i$ based on feature similarity to get the predicted coarse correspondence set $\hat{C}_{coarse} = \{(P_i, u_i)\}$, we use cosine similarity $s(\cdot,\cdot)$ to denote the similarity between two features:
\begin{equation}
    s(P_i, I_j) = \frac{<{F}^{coarse}_{P_i},{F}^{coarse}_{I_j}>}{\|{F}^{coarse}_{P_i}\|\|{F}^{coarse}_{I_j}\|}
\end{equation}



\noindent
\textbf{Inter-intra Contrastive Learning (ICL).} 
Unlike normal contrastive learning (NCL) in~\cite{kang2023cofii2p, yao2023cfi2p} that solely considers inter-modal pairs, our ICL further incorporates intra-modal pairs (see Figure~\ref{fig:coarse_matching} Similarity Matrix). By constructing negative pairs within the same modality (e.g., $I_1$ and $I_2$), the feature distance between different image patches is increased, as well as between different point groups.
Therefore, ICL is not only to bring the features of 3D point groups and their corresponding image patches closer together but also to ensure that the 2D patch features within each image and 3D point groups within each point cloud remain as differentiated as possible. This balance enhances more precise alignment while preserving distinctiveness among local intra-features. Similarity visualizations are in Sec.~\ref{sec: ablation_study}.

Our ICL loss is defined as: 
\begin{equation}
    L^{S}_{coarse} = \log[1+\sum_{j}\exp^{\alpha_{p}(1-s^j_{p}+m_{p})} \sum_{k}\exp^{\alpha_{n}(s^k_{n}-m_{n})}]
\label{equation: inter-intra}
\end{equation}
where $s_{p}^{j}$ and $s_{n}^{k}$ denote the cosine similarity of positive and negative pairs, respectively.
The positive pairs are determined by whether the center of point group $P_i$ is projected into the image patch $I_i$.
$\alpha_{p}$ and $\alpha_{n}$ are the adaptive weighting factors for positive and negative pairs, respectively:
\begin{equation}
    \begin{aligned}
        & \alpha_p = \gamma \cdot \max(0, 1+m_p-s^j_p) \\ 
        & \alpha_n = \gamma \cdot \max(0, s^k_n-m_n)
    \end{aligned}
\end{equation}
in which $\gamma$ is the scale factor and $m_{p}$ and $m_{n}$ are positive and negative margin for better similarity separation.

\noindent
\textbf{Dense Training Alignment (DTA).}
Moreover, we find that the proposed ICL can only align features with sparsely sampled image patch-point groups, neglecting additional global features. Inspired by ~\cite{zhang2024telling}, 
we propose a dense training
alignment strategy to back-propagate the calculated gradients to all image patches using a soft-argmax operator.
Specifically, for each sampled point group feature $P_x$, we compute its similarity map $S_x={F}^{coarse}_{P_x} {{F}^{coarse}_{I}}^{\mathrm{T}} \in \mathbb{R}^{1 \times H/s \times W/s}$ with the target image feature ${F}^{coarse}_{I}$, as shown in Figure~\ref{fig:coarse_matching}. Then, we take soft-argmax over the similarity map to compute the predicted pixel position $\hat{u}_x=\text{SoftArgmax}(S_x)$. 
We penalize the distance between the predicted position $\hat{u}_x$ and the target position $u_x$ with a L2 norm loss, which is defined as: 
\begin{equation}
    L^{D}_{coarse} = \sum^{\kappa}_{x} \|\hat{u}_x-u_x\|_{2},
\end{equation}


\subsection{Fine Matching}
\label{sec:fm}
After the coarse matching, we aim to generate point-to-pixel pairs in fine matching.
We first generate the image feature $\mathbf{F}^{Fine}_{I} \in \mathbb{R}^{H/2 \times W/2 \times C^{\prime}}$ and the point feature $\mathbf{F}^{Fine}_{P} \in \mathbb{R}^{N \times C^{\prime}}$ in fine-resolution via applying the two upsample networks, ResNet~\cite{he2016deep} and PointNet~\cite{pointnet}, for image and point cloud respectively.
We then extract the local feature from the fine-resolution image patch, $\mathbf{F}^{Fine}_{I_x} \in \mathbb{R}^{w \times w \times C^{\prime}}$, centered on the predicted coarse pixel coordinate $\hat{u}_x$.
Additionally, we extract the corresponding feature $\mathbf{F}^{Fine}_{P_x} \in \mathbb{R}^{1 \times C^{\prime}}$ of the center point $P_x$ within the point group.  
The fine matching process is defined as:
\begin{equation}
\begin{aligned}
    & \hat{S}^{Fine}_{x} = \mathbf{F}^{Fine}_{P_x} {\mathbf{F}^{Fine}_{I_x}}^{\mathrm{T}} \in \mathbb{R}^{1 \times w \times w} \\
    & \hat{u}^{Fine}_{x} = \text{SoftArgmax}(\hat{S}^{Fine}_{x})
\end{aligned}
\end{equation}
where $\hat{S}^{Fine}_{x}$ is the fine similarity map between the point center and extracted image patch, and $\hat{u}^{Fine}_{x}$ is the final predicted 2D pixel corresponding to 3D point $P_x$. 


\subsection{Pose Estimation}
\label{sec:pe}
Following~\cite{kang2023cofii2p}, we use the  EPnP-RANSAC~\cite{epnp,pnpsolver} algorithm to filter out incorrect pixel-point pairs and estimate the relative pose of the camera based on the set of predicted pixel-point correspondence after the fine matching.

\subsection{Loss Function}
\label{sec: loss}
Our training loss includes four components: 

\noindent
\textbf{Geometry-guided Attention Loss.} We use a Binary Cross Entropy (BCE) loss $L_{Att}$ to supervise the cross-attention layers at the final stage of the Fusion Transformer during cross-modal feature fusion (See Sec.~\ref{sec: geocros}).

\noindent
\textbf{In-frustum Detection Loss.}
We use a standard binary cross-entropy (BCE) loss $L_{det}$ (See Fig.~\ref{fig:coarse_matching}) to supervise the in-frustum super-points classification.

\noindent
\textbf{Coarse Matching Loss.} 
In coarse matching, 
we first sample $\kappa$ pairs from the ground-truth 2D-3D corresponding set $C^{*}_{coarse}= \{(P_x, I_{x})|I_{x}=\mathcal{F}(T_{GT}P_x)\}$,
where $T_{GT}$ is the ground-truth transformation matrix from point cloud coordinate system to image frustum coordinate system, and $\mathcal{F}$ denotes the mapping function that convert points from camera frustum to image patch position. We determine the negative pairs by checking whether the distance between the location of $I_{x}$ and the projection of $P_x$ on the image is larger than a threshold $r$ or not. 
The process is the same within each individual modality. Then, we combine the Inter-intra Contrastive Learning loss and the loss in Dense Training Alignment to form our coarse matching loss $L_{coarse} = L^{S}_{coarse} + L^{D}_{coarse}$.

\noindent
\textbf{Fine Matching Loss.} Since the fine image patch has a small region, we utilize Cross Entropy (CE) loss $L^{S}_{fine}$ to sparsely supervise the fine matching process and apply dense L2 norm loss $L^{D}_{fine}$ for constructing fine matching loss $L_{fine} = L^{S}_{fine} + L^{D}_{fine}$:
\begin{equation}
\begin{aligned}
    & L^{S}_{fine} = - \frac{1}{\kappa} \sum_{x=1}^\kappa \sum_{c=1}^{w^2} S^{Fine}_{x,c} \log(softmax(\hat{S}^{Fine}_{x,c}))\\
    & L^{D}_{fine} = \sum^{\kappa}_{x} \|\hat{u}^{Fine}_{x}-u^{Fine}_{x}\|_{2}
\end{aligned}
\end{equation}
where similarity value $S^{Fine}_{x,c}$ equals to 1 when the 3D point $P_x$ is corresponded with the $c_{th}$ pixel in the flatten image patch, else 0. $\kappa$ is the number of sampled 2D-3D pairs, $w$ is the side length of image patch and ${u}^{Fine}_{x}$ denotes the GT corresponded 2D pixel. To avoid overfitting, we randomly shift the centered position of the extracted fine patch by up to $\pm w/2$ pixels, preventing the ground-truth pixel from always being at the center of the patch.

Overall, our loss function is:
\begin{equation}
    L = \lambda_1 L_{Att} + \lambda_2 L_{det} + \lambda_3 L_{coarse} + \lambda_4 L_{fine}
\end{equation}
where $\lambda_1,\lambda_2,\lambda_3,\lambda_4$ regulate the losses' contributions.

\section{Experiments}

\begin{table*}[h]
\centering
\resizebox{\linewidth}{!}{
\begin{tabular}{c|c|c|c|c|c|c|c|c|c|c|c|c|c}
\hline
\multirow{2}{*}{} & \multicolumn{2}{c|}{$\textbf{Test}_{T1-T7}$} & \multicolumn{2}{c|}{$\textbf{Test}_{T1-T7 hard}$} & \multicolumn{2}{c|}{$\textbf{Test}_{T10}$} & \multicolumn{3}{c|}{KITTI Odometry} & \multicolumn{3}{c|}{Nuscenes} & Runtime \\ \cline{2-14}
\multirow{2}{*}{} & RRE($^\circ$) & RTE($m$) & RRE($^\circ$) & RTE($m$) & RRE($^\circ$) & RTE($m$) & RRE($^\circ$) & RTE($m$) & RR(\%) & RRE($^\circ$) & RTE($m$) & RR(\%) &$t$(sec) \\ \hline
DeepI2P-Cls~\cite{deepi2p}      & 9.02  & 6.31   & 9.10   & 6.12    & 18.30   & 11.46  & 5.88  & 1.13   & 80.18   & 7.37    & 2.22   & 62.67 & \textbf{0.23} \\ 
DeepI2P-2D~\cite{deepi2p}       & 20.31  & 7.01   & 33.43   & 7.93    & 49.93   & 17.09  & 3.87  & 1.42   & 74.50   & 3.02    & 1.95   & 92.53 & 8.78  \\ 
DeepI2P-3D~\cite{deepi2p}       & 17.97  & 7.29   & 34.86   & 9.49    & 43.11   & 17.41  & 6.08  & 1.21   & 38.34   & 7.06    & 1.73   & 18.82 & 18.62  \\ 
CorrI2P~\cite{ren2022corri2p}          & 20.08  & 12.63   & 27.95   & 13.97    & 32.23  & 14.45  & 2.72  & 0.90   & 92.19   & 2.31    & 1.7   & 93.87 & 1.33  \\ 
VP2P~\cite{zhou2024differentiable}   & / & / & / & / & / & / & 2.39  & 0.59   & 95.07   & 2.15    & 0.89   & /  & 0.76 \\ 
CFI2P~\cite{yao2023cfi2p}            & 4.46  & \underline{1.92}   & 8.56   & \underline{2.48}    & \underline{10.81}   & \underline{7.15}  & 1.38  & 0.54   & 99.44   & \underline{1.47}    & 1.09   & \underline{99.23} & \underline{0.33}  \\ 
CoFiI2P~\cite{kang2023cofii2p}          & \underline{4.24}  & 2.82   & \underline{7.87}   & 5.34    & 17.78   &7.43  & \underline{1.14}  & \underline{0.29}   & \textbf{100.00}   & 1.48    & \underline{0.87}   & 98.67 & 0.64  \\ \hline
Ours with $K_{Pred}$ & 2.04  & 1.72    & 4.56   & 2.19    & 4.61   & 4.80 & / & / & / & / & / & / & 1.74 \\
Ours with $K_{GT}$ & \textbf{0.66}  & \textbf{0.51}    & \textbf{2.64}   & \textbf{1.13}    & \textbf{2.53}   & \textbf{2.69}  & \textbf{0.87}  & \textbf{0.19}    & \textbf{100.00}   & \textbf{1.38}    & \textbf{0.78}   & \textbf{99.45} & 0.85  \\ \hline
\end{tabular}
}
\caption{Quantitative localization results on the proposed \textit{Carla Intersection}, \textit{KITTI~\cite{kitti}} and \textit{Nuscenes~\cite{NuScenes}} datasets. We report median RRE and median RTE for \textit{Carla Intersection} and mean RRE and mean RTE for \textit{KITTI} and \textit{Nuscenes} following previous Image-to-Point cloud registration methods~\cite{kang2023cofii2p, ren2022corri2p, deepi2p}.
Our model achieves the best performance on all datasets, especially on unseen scenarios. 
}
\label{tab:result_carla}
\vspace{-0.3cm}
\end{table*}

\subsection{Proposed Carla Intersection Dataset}

The proposed \textit{Carla Intersection} dataset comprises 75 intersections across 8 worlds within the Carla~\cite{Carla} simulation environment, encompassing urban and rural landscapes. We use on-board LiDAR sensor to capture point cloud scans, which are then accumulated and downsampled to get the 3D point cloud of the intersection. For each intersection, we captured 768 training images and 288 testing images with known 6-DoF pose at a resolution of 1920x1080 pixel and a horizontal field of view (FOV) of $90^\circ$. In consideration of real-world traffic surveillance camera installations, our image collection spans heights from 6 to 8 meters, with camera pitch angles from 15 to 30 degrees. This setup reflects typical positioning to capture optimal traffic views under varied monitoring conditions. 
More details and visualization of our dataset are in the Supplementary Materials.

To access the model's generalization capability, we trained on 67 intersections from worlds $Town01$ to $Town07$, and tested on one unseen intersection from each of the 8 worlds. We divided the testing data into three sets:

$\textbf{Test}_{T1-T7}$ contains 7 unseen intersections from world $Town01$ to $Town07$, with 288 testing images each, matching the training pitch angles (15 or 30 degree).

$\textbf{Test}_{T1-T7hard}$ contains 7 same intersection scenes as in $\textbf{Test}_{T1-T7}$, with images at pitch angles of 20 or 25 degrees, to evaluate robustness against viewpoint variations.

$\textbf{Test}_{T10}$ contains 1 unseen intersection scene from unseen world $Town10$. This split sets a high standard for evaluating the model's generalization ability, assessing its performance in unseen urban styles and intersections.



\subsection{Experimental Setup}

\textbf{Datasets.} We conduct experiments on the \textit{Carla Intersection} dataset. We also evaluate TrafficLoc on the real USTC intersection dataset~\cite{sheng2024rendering}, and two in-vehicle camera benchmarks, KITTI Odometry~\cite{kitti} and Nuscenes~\cite{NuScenes}. For fair comparisons, we use the same training and evaluation pairs of image and point cloud data on KITTI and Nuscenes following previous I2P registration methods~\cite{ren2022corri2p,deepi2p}. 

\noindent
\textbf{Implementation Details.} The training details and others are in the Supplementary Materials.
\noindent
\textbf{Evaluation Metrics.} Following previous works~\cite{kang2023cofii2p, ren2022corri2p, deepi2p}, we evaluate the localization performance with relative rotation error (RRE), relative translation error (RTE) and registration recall (RR). RRE and RTE are defined as:
\begin{equation}
\vspace{-0.1cm}
    \mathrm{RRE}=\sum_{i=1}^{3}|\mathbf{r}(i)|,\quad \mathrm{RTE}=||\mathbf{t}_{gt}-\mathbf{t}_{pred}||
\vspace{-0.1cm}
\end{equation}
where $\mathbf{r}$ is the Euler angle vector of $\mathbf{R}^{-1}_{gt}\mathbf{R}_{pred}$, $\mathbf{R}_{gt}$ and $\mathbf{t}_{gt}$ are the ground-truth rotation and translation matrix, $\mathbf{R}_{pred}$ and $\mathbf{t}_{pred}$ represent the estimated rotation and translation matrix. RR denotes the fraction of successful registrations among the test dataset. 
A registration is considered as successful when the RRE is smaller than $\tau_{r}=10^{\circ}$ and the RTE is smaller than $\tau_{t}=5m$.

\begin{figure}
    \centering
    \includegraphics[width=0.9\linewidth]{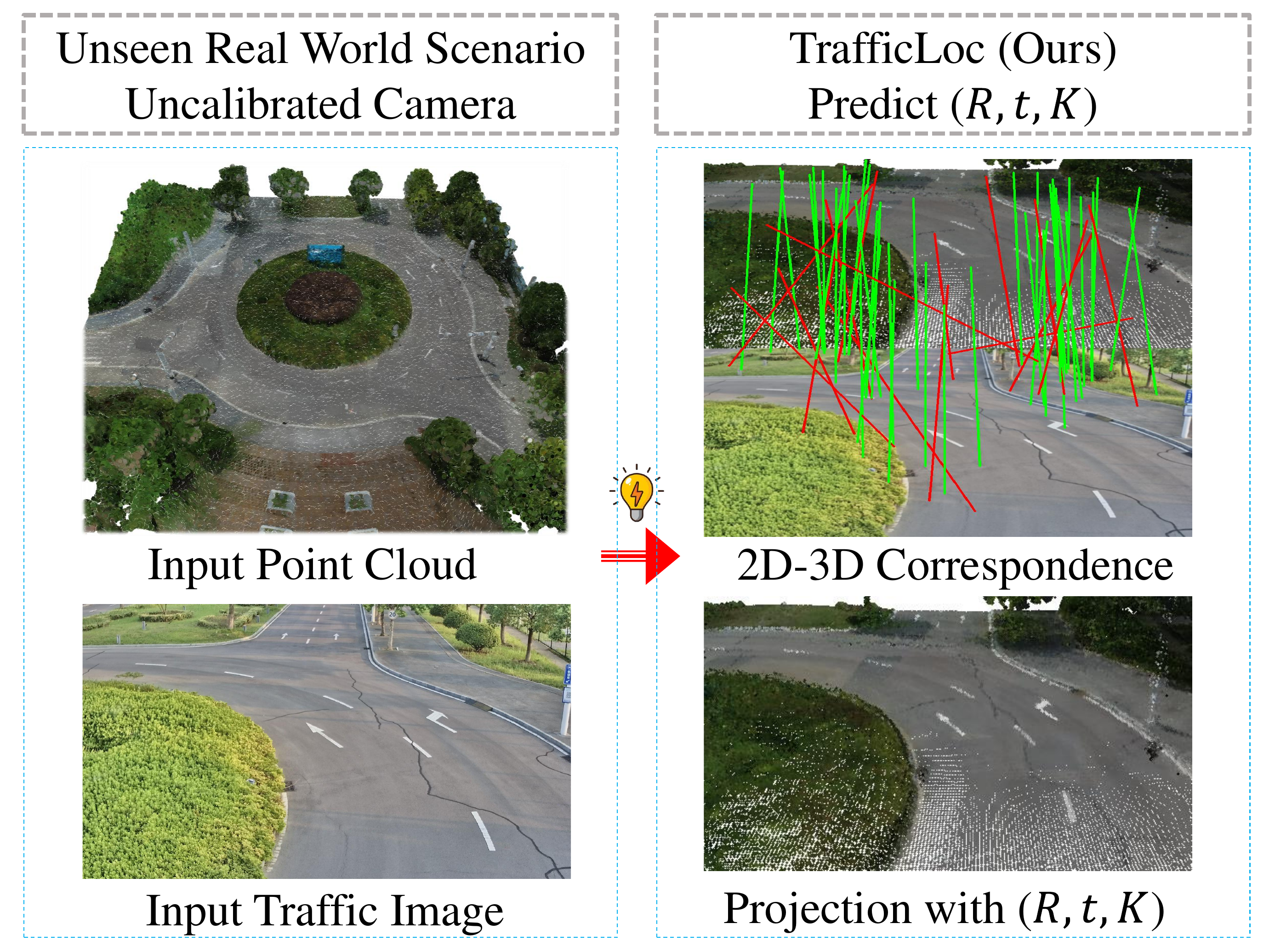}
    \caption{Localization performance of our TrafficLoc on the USTC intersection dataset~\cite{sheng2024rendering}. Note that the model is trained on the synthetic \textit{Carla Intersection} dataset.}
    \label{fig:ustc_result}
\vspace{-0.6cm}
\end{figure}

\subsection{Evaluation Results}
We first evaluate TrafficLoc on our \textit{Carla Intersection} dataset and compare it with other baseline methods. Table \ref{tab:result_carla} summarizes the results.
Our method outperforms all baseline methods by a large margin in all three test splits. This indicates that TrafficLoc is robust to large viewpoint changes and has great generalization ability on \textbf{unseen} traffic scenarios. Specifically, the RRE reduces \textbf{85\%}, \textbf{66\%}, \textbf{86\%} and the RTE reduces \textbf{82\%}, \textbf{78\%}, \textbf{64\%} compared to the previous state-of-the-art CoFiI2P~\cite{kang2023cofii2p} in three test splits respectively. 
Even in entirely unseen city-style environments and unseen scenes ($\textbf{Test}_{T10}$), our model maintains robust localization capability, while other baseline methods fail. Additionally, our model achieves high accuracy while maintaining efficient inference time. 
When intrinsic parameters $K$ are unknown, we use intrinsics predicted by DUSt3R~\cite{wang2024dust3r} to initialize the intrinsics for EPnP-RANSAC~\cite{epnp,pnpsolver}, which slightly decreases performance and inference speed but still yields strong results. 
The result of VP2P~\cite{zhou2024differentiable} on \textit{Carla Intersection} dataset and its RR on Nuscenes are unavailable since their code is not provided.

Besides \textit{Carla Intersection} dataset, we evaluate our TrafficLoc on KITTI and Nuscenes benchmarks.
TrafficLoc achieves the best performance on both datasets, achieving \textbf{34\%} RTE improvement compared to the previous state-of-the-art CoFiI2P~\cite{kang2023cofii2p} and $<1^\circ$ RRE first-time on KITTI, indicating its strong ability also on in-vehicle view cases.

To test the Sim2Real generalizability of our TrafficLoc, we evaluate it on the real-world intersection from the USTC dataset~\cite{sheng2024rendering}, after training on the \textit{Carla Intersection}. Note that the test intersection is totally unseen and the traffic camera is uncalibrated. 
Figure~\ref{fig:ustc_result} shows the qualitative localization result. Since ground-truth pose is unavailable, we project the point cloud onto the image plane with the predicted transformation matrix
$[{R}|{t}]$ 
and intrinsic parameters $K$. 
The projection image shows a clear overlap with the input image, validating the accuracy of our method.

\subsection{Ablation Study}
~\label{sec: ablation_study}
We evaluate the effectiveness of different proposed components in our TrafficLoc on the \textit{Carla Intersection}
dataset $\textbf{Test}_{T1-T7}$ here. More ablation studies and discussions are given in the \textbf{Supplementary Materials}.

\begin{figure}
    \centering
    \includegraphics[width=1.0\linewidth]{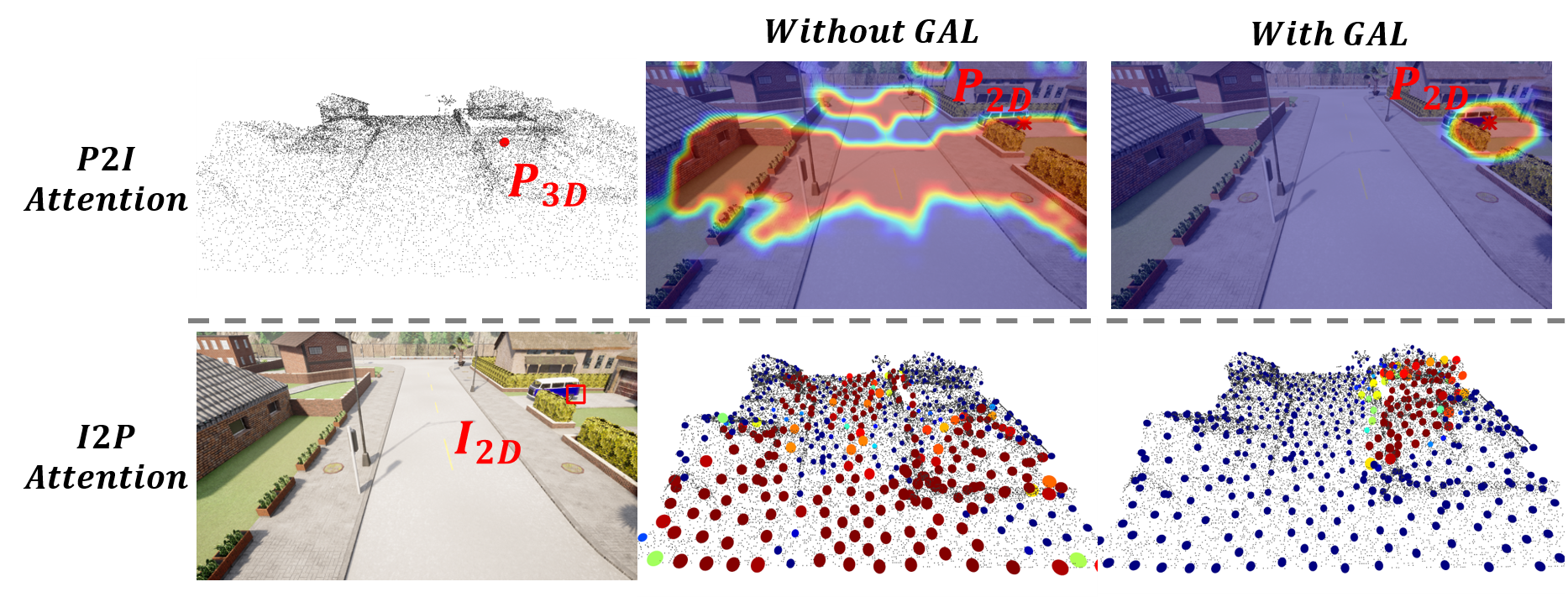}
    \caption{Visualization result of P2I and I2P attention map when using Geometry-guided Attention Loss $L_{att}$ or not. Red color indicates high attention value and blue means low value.}
    \label{fig:ablation_att}
\vspace{-0.2cm}
\end{figure}

\noindent
\textbf{Geometry-guided Attention Loss (GAL).} Fig.~\ref{fig:ablation_att}  highlights the advanced geometry-awareness ability of the proposed GAL. With the use of GAL, the P2I attention map of point group $P_{3D}$ tends to concentrate more on the image region where the point group is projected, while the I2P attention map for patch $I_{2D}$ assigns greater weights to the area traversed by the camera ray of this patch. As shown in Table \ref{tab:ablation_matching} (last two rows), with the proposed $L_{att}$, our model brings notable improvements (\textbf{17.7\%}) in the RTE metric.

\begin{figure}
    \centering
    \includegraphics[width=1.0\linewidth]{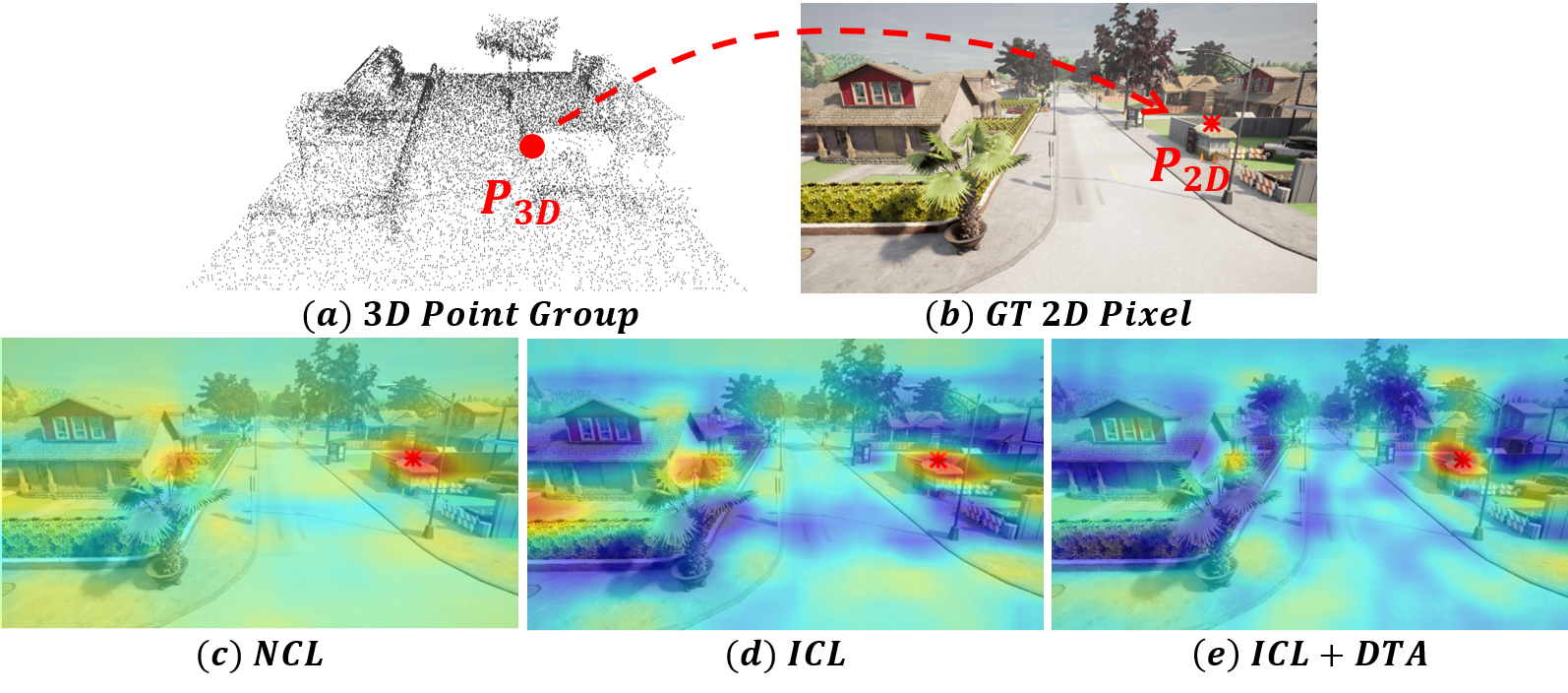}
    \caption{Visualization result of using different loss function. (a), (b) denote the point group center $P_{3D}$ and its corresponding pixel $P_{2D}$. (c), (d), (e) show the similarity map between point group and image feature. Blue means low similarity and red means high.}
    \label{fig:ablation_loss}
    \vspace{-0.3cm}
\end{figure}

\noindent
\textbf{Inter-intra Contrastive Learning (ICL).}
Fig.~\ref{fig:ablation_loss} (c) and (d) vividly show the advantages of the proposed ICL. When using existing normal contrastive loss (NCL), the similarity map exhibits generally high values throughout. With the proposed ICL, the distinction within the similarity map increases, demonstrating our TrafficLoc can achieve a more pronounced distributional difference between features within the same modality. Quantitative results in Table \ref{tab:ablation_matching} (first two rows) demonstrate we achieve \textbf{9.8\%} improvements in RTE with the proposed ICL compared to using previous NCL. 

\noindent
\textbf{Dense Training Alignment (DTA).} From Fig.~\ref{fig:ablation_loss} (d), we can see that multiple peaks (red regions) still remain due to the sparse supervision during training. With the proposed DTA, the similarity map in Fig.~\ref{fig:ablation_loss} (e) only has a single peak region, showing strong robustness in feature matching. Table \ref{tab:ablation_matching} (second and third rows) sees an improvement of \textbf{20.5\%} and \textbf{16.2\%} in RRE and RTE after using the proposed DTA. This is due to our DTA having global supervision across the image patch-point group matching during training.

\vspace{-0.1cm}
\vspace{-0.1cm}

\begin{table}[h]
\centering
\resizebox{\columnwidth}{!}{
\begin{tabular}{c|c|c|c|c|c}
    & $CM$ & $FM$ & GAL & RRE ($^\circ$) & RTE ($m$) \\ \hline
\multirow{5}{*}{Baseline} & NCL  & /   & /   & 1.53    & 0.82   \\ 
    & ICL  & /   & /   & 1.27    & 0.74   \\ 
    & ICL + DTA  & /   & /   & 1.01    & 0.62   \\ 
    & ICL + DTA  & \checkmark   & /   & 0.84    & 0.62   \\ \hline
Ours   & ICL + DTA  & \checkmark    & \checkmark   & \textbf{0.66}   & \textbf{0.51}  \\ 
\end{tabular}
}
\caption{Ablation Study of our TrafficLoc on the Carla Intersection $\textbf{Test}_{T1-T7}$. $CM$ denotes Coarse Matching and $FM$ denotes Fine Matching. 
``GAL'' indicates the proposed Geometry-guided Attention Loss.
``NCL'' means using normal contrastive learning, while ``ICL'' means using the proposed Inter-intra Contrastive Learning. ``DTA'' indicates using the proposed Dense Training Alignment.  
}
\label{tab:ablation_matching}
\vspace{-0.2cm}
\end{table}

\section{Conclusion}
In this work we focus on the under-explored problem of traffic camera localization, which is an important capability for fully-integrated spatial awareness among city-scale camera networks and vehicles.
Such large-scale sensor fusion has the potential to enable more robustness, going beyond the limitations of a single vehicle's point of view.
We proposed a novel method, TrafficLoc, which 
we show to be effective.
To facilitate training and evaluation we propose the novel \textit{Carla Intersection} dataset, focusing on the case of intersections, which is a common placement for traffic cameras and a focal point for traffic safety.
We hope that this dataset will facilitate more research into integrated camera networks for robust, cooperative perception.

{
    \small
    \bibliographystyle{ieeenat_fullname}
    \bibliography{main}
}

\clearpage
\setcounter{page}{1}
\maketitlesupplementary

\renewcommand{\thesection}{\Alph{section}}
\setcounter{section}{0}

\begin{table*}[t]
\centering
\begin{tabular}{c|c|c|c|c|c|c|c|c|c}
    & \multirow{2}{*}{$CM$} & \multirow{2}{*}{$FM$} & \multirow{2}{*}{GAL} & \multicolumn{2}{c|}{$\textbf{Test}_{T1-T7}$} & \multicolumn{2}{c|}{$\textbf{Test}_{T1-T7 hard}$} & \multicolumn{2}{c}{$\textbf{Test}_{T10}$} \\ \cline{5-10}
    &  &  &  & RRE($^\circ$) & RTE($m$) & RRE($^\circ$) & RTE($m$) & RRE($^\circ$) & RTE($m$) \\ \hline
\multirow{5}{*}{Baseline} & NCL  & /   & /   & 1.53    & 0.82 & 3.79 & 1.98 & 7.35 & 7.47  \\ 
    & ICL  & /   & /            & 1.27    & 0.74 & 3.72 & 1.87 & 4.09 & 3.26   \\ 
    & ICL  & /   & \checkmark   & 0.95    & 0.64 & 3.12 & 1.46 & 3.03 & 2.86   \\ 
    & ICL + DTA  & /   & /        & 1.01    & 0.62 & 3.23 & 1.65 & 2.99 &2.80   \\ 
    & ICL + DTA  & \checkmark     & /   & 0.84    & 0.62 & 3.17 & 1.42 & 2.98 & 2.83   \\ \hline
Ours   & ICL + DTA  & \checkmark    & \checkmark   & \textbf{0.66}   & \textbf{0.51} & \textbf{2.64} & \textbf{1.13} & \textbf{2.53} & \textbf{2.69}  \\ 
\end{tabular}
\caption{Ablation Study on loss function and model design. We report median RRE and median RTE results on all three test splits of \textit{Carla Intersection} dataset. $CM$ denotes Coarse Matching and $FM$ denotes Fine Matching. ``NCL'' means using normal contrastive learning, while ``ICL'' means using the proposed Inter-intra Contrastive Learning. ``DTA'' indicates using the proposed Dense Training Alignment. ``GAL'' represents applying the proposed Geometry-guided Attention Loss.}
\label{tab:supp_ablation_loss}
\vspace{-0.2cm}
\end{table*}

\section{Overview}
In this supplementary material, we provide a detailed explanations of our TrafficLoc and the proposed \textit{Carla Intersection} dataset. Additionally,  we present extended experimental results on the \textit{Carla Intersection} dataset and KITTI Odometry dataset~\cite{kitti}, showcasing the robust localization capabilities of our TrafficLoc and offering further insights we gathered during the development.

We begin by presenting more experimental results and comprehensive ablation studies and analysis in Sec. \ref{sec: supp_ablation}. 
In Sec. \ref{sec: supp_carla_dataset}, we outline the data collection process and provide visualizations of our \textit{Carla Intersection} dataset. 
Sec. \ref{sec: supp_gff} describes the detailed elements of the Fusion Transformer in the GFF module, followed by Sec. \ref{sec: supp_implement} with implementation details of our network architecture and training procedure. 
Finally, Sec. \ref{sec: supp_visual} offers additional visualizations of our localization results across different datasets.

\begin{table*}[t]
\centering
\resizebox{\linewidth}{!}{
\begin{tabular}{c|c|c|c|c|c|c|c|c|c|c|c}
    & \multirow{2}{*}{$\theta_{low}(^\circ)$} & \multirow{2}{*}{$\theta_{up}(^\circ)$} & \multirow{2}{*}{$d_{low}(m)$} & \multirow{2}{*}{$d_{up}(m)$} & \multirow{2}{*}{Layer} & \multicolumn{2}{c|}{$\textbf{Test}_{T1-T7}$} & \multicolumn{2}{c|}{$\textbf{Test}_{T1-T7 hard}$} & \multicolumn{2}{c}{$\textbf{Test}_{T10}$} \\ \cline{7-12}
    &  &  &  & & & RRE($^\circ$) & RTE($m$) & RRE($^\circ$) & RTE($m$) & RRE($^\circ$) & RTE($m$) \\ \hline
\multirow{7}{*}{Baseline} & /  & /   & /   & /    & / & 0.84 & 0.62 & 3.17 & 1.42 & 2.98 & 2.83 \\ 
    & 10  & 10   & 3   & 5    & Last & 1.24 & 0.80 & 3.49 & 1.53 & 6.07 & 7.45   \\ 
    & 10  & 20   & 3   & 3    & Last & 1.27 & 0.83 & 3.05 & 1.46 & 3.55 & 2.95  \\ 
    & 20  & 30   & 3   & 5    & Last & 0.91 & 0.59 & 2.71 & 1.27 & 3.05 & 2.78  \\ 
    & 10  & 20   & 5   & 7    & Last & 0.85 & 0.55 & 2.63 & 1.15 & 3.08 & 2.75  \\ 
    & 10  & 20   & 3   & 5    & First &1.00 & 0.59 & 2.68 & 1.14 & 4.30 & 3.23 \\ 
    & 10  & 20   & 3   & 5    & All  & 1.02 & 0.62 & 3.01 & 1.19 & 3.45 & 3.33 \\ \hline
Ours& 10  & 20   & 3   & 5    & Last &\textbf{0.66} &\textbf{0.51} &\textbf{2.64} &\textbf{1.13} &\textbf{2.53} &\textbf{2.69} \\ 
\end{tabular}
}
\caption{Ablation Study on Geometry-guided Attention Loss (GAL). $\theta_{low}$ and $\theta_{up}$ denote the angular threshold for image-to-point cloud (\textbf{I2P}) attention, while $d_{low}$ and $d_{up}$ represent the distance threshold for point cloud-to-image (\textbf{P2I}) attention. ``Layer'' specifies the fusion layer within the Geometry-guided Feature Fusion (GFF) module where GAL is applied.}
\label{tab:supp_gal}
\vspace{-0.2cm}
\end{table*}

\section{More Ablation Studies and Analysis}
\label{sec: supp_ablation}
In this section, we show more experimental results and ablation studies to evaluate the effectiveness of different proposed components in our TrafficLoc.

\noindent
\textbf{Geometry-guided Attention Loss (GAL).} As shown in Table~\ref{tab:supp_ablation_loss}, the model incorporating GAL consistently outperforms its counterpart without GAL across all three test splits of the \textit{Carla Intersection} dataset. Notably, when evaluated on images with unseen pitch angles from the $\textbf{Test}_{T1-T7 hard}$ split, TrafficLoc achieves remarkable \textbf{20.4\%} improvement in RTE (see the last two rows), highlighting the robustness of GAL to viewpoint variations.

\noindent
\textbf{Inter-intra Contrastive Learning (ICL).} The first and second rows of Table~\ref{tab:supp_ablation_loss} present the ablation study results comparing ICL and normal contrastive learning (NCL). When using NCL in coarse matching, the model exhibits relatively high error across all three test splits, particularly on $\textbf{Test}_{T10}$ which features an unseen world style. Leveraging  ICL significantly improves performance, achieving \textbf{44.3\%} and \textbf{56.3\%} gains in RRE and RTE on $\textbf{Test}_{T10}$, respectively. This enhancement brings the localization accuracy in unseen scenes from an unseen world style to a reasonable level, making reliable localization feasible.

\noindent
\textbf{Dense Training Alignment (DTA).} The second and fourth rows of Table~\ref{tab:supp_ablation_loss} present the ablation study results of DTA, which facilitates global image supervision by allowing gradients to back-propagate through all image patches via soft-argmax operation. With the proposed DTA, we observe an improvement of \textbf{26.9\%} and \textbf{14.1\%} in RRE and RTE, respectively, on $\textbf{Test}_{T10}$.

\noindent
\textbf{Fine Matching (\textit{FM}).} The fine matching module refines point-to-pixel correspondences within the point group–image patch pairs derived from the coarse matching results. As shown in the fourth and fifth rows of Table \ref{tab:supp_ablation_loss}, the fine matching module further enhances the model's localization accuracy in seen world styles, achieving a \textbf{16.8\%} improvement in RRE on the $\textbf{Test}_{T1-T7}$ split.



\noindent
\textbf{More analysis of GAL.} The ablation results for the Geometry-guided Attention Loss (GAL) are summarized in Table~\ref{tab:supp_gal}. We conducted experiments on the \textit{Carla Intersection} dataset with GAL using different threshold parameters 
and applying GAL across different layers of the Geometry-guided Feature Fusion (GFF) module. 

When the lower and upper threshold are set to the same value (see the second and third row), the model performs worse than not applying GAL, which highlights the importance of defining a tolerant region that enables the network to flexibly learn attention relationships for intermediate cases between the lower and upper thresholds. 
With thresholds $\theta_{low}$, $\theta_{up}$, $d_{low}$ and $d_{up}$ set to 10$^\circ$, 20$^\circ$, \SI{3}{m}
and \SI{5}{m}, our model consistently outperforms the baseline without GAL across all metrics.
Moreover, we observed that applying GAL to either the first layer or all layers of the GFF module yields worse localization results compared to applying it only to the last layer. This is mostly because such configurations constrain the network's ability to capture global features during the early stages (or initial layers) of multimodal feature fusion.

\begin{table}[h]
\centering
\resizebox{\columnwidth}{!}{
\begin{tabular}{c|c|c|c|c|c}
Base  & \multirow{2}{*}{DTA} & \multirow{2}{*}{GAL} & \multirow{2}{*}{RRE($^\circ$)} & \multirow{2}{*}{RTE($m$)} & \multirow{2}{*}{RR(\%)} \\
Model & & & & &  \\ \hline
CoFiI2P   & $\times$  & $\times$    & 1.14 & 0.29 & \textbf{100.00}   \\ 
CoFiI2P   & \checkmark  & $\times$   & 0.94 & 0.24 & \textbf{100.00}   \\ 
CoFiI2P   & $\times$  & \checkmark   & 1.01 & 0.27 & \textbf{100.00}  \\ 
CoFiI2P   & \checkmark  & \checkmark   & \textbf{0.85} & \textbf{0.22} & \textbf{100.00}  \\ 
\end{tabular}
}
\caption{Experimental results on KITTI Odometry dataset~\cite{kitti} based on current SOTA model CoFiI2P~\cite{kang2023cofii2p}. ``DTA'' and ``GAL'' mean whether we add the proposed Dense Training Alignment and Geometry-guided Attention Loss into CoFiI2P during the training, respectively. We report the mean RRE, mean RTE, and RR metrics for comparison.}
\label{tab:supp_ablation_cofii2p}
\vspace{-0.2cm}
\end{table}

\noindent
\textbf{More analysis of DTA and GAL.} To further verify the effectiveness of our Dense Training Alignment (DTA) and Geometry-guided Attention Loss (GAL) in enhancing previous Image-to-Point Cloud (I2P) registration methods, we integrated them into the state-of-the-art CoFiI2P network \cite{kang2023cofii2p}. 
We conducted experiments on the KITTI Odometry dataset \cite{kitti} to assess the improvements. 
The experimental results shown in Table~\ref{tab:supp_ablation_cofii2p} demonstrate that CoFiI2P achieves improved performance on both RRE and RTE metrics when equipped with DTA or GAL. Notably, with both components applied, CoFiI2P achieves improvements of \textbf{25.4\%} and \textbf{24.1\%} in RRE and RTE, respectively.

\begin{table}[h]
\centering
\resizebox{\columnwidth}{!}{
\begin{tabular}{c|c|c|c|c}
    & Img Encoder & PC Encoder & RRE($^\circ$) & RTE($m$) \\ \hline
\multirow{4}{*}{Baseline}    & ResNet~\cite{he2016deep}  & PiMAE~\cite{chen2023pimae}      & 1.25    & 0.87   \\ 
& ResNet~\cite{he2016deep}  & PT~\cite{point_transformer}       & 0.85    & 0.58   \\ 
    & DUSt3R~\cite{wang2024dust3r}  & PiMAE~\cite{chen2023pimae}     & 1.03    & 0.75   \\ 
    & DUSt3R$^{\ast}$~\cite{wang2024dust3r}  & PT~\cite{point_transformer}     & 0.77    & 0.59   \\ \hline
Ours   & DUSt3R~\cite{wang2024dust3r}  & PT~\cite{point_transformer}      & \textbf{0.66}   & \textbf{0.51}  \\ 
\end{tabular}
}
\caption{Ablation study on feature extraction backbone. We report median RRE and median RTE results on test split $\textbf{Test}_{T1-T7}$. DUSt3R$^{\ast}$ means using frozen DUSt3R backbone during training.}
\label{tab:ablation_backbone}
\vspace{-0.3cm}
\end{table}

\noindent
\textbf{Feature extraction backbone.} Table \ref{tab:ablation_backbone} illustrates the results under different image and point cloud feature extraction backbone. Our model performs best when using DUSt3R~\cite{wang2024dust3r} and Point Transformer~\cite{point_transformer} as backbones, benefiting from DUSt3R's strong generalization ability. Even with a frozen DUSt3R, the model achieves comparable performance. In contrast, when using ResNet~\cite{he2016deep} or PiMAE~\cite{chen2023pimae}, the model's performance declines due to the lack of attentive feature aggregation during the feature extraction stage. When utilizing PiMAE, we load the pretrained weights of its point encoder.

\noindent
\textbf{Localization with unknown intrinsic parameters.} Ablation results of localization with predicted intrinsic parameters are shown in Table~\ref{tab:supp_intrinsic}. In the absence of ground-truth intrinsic parameters during inference, we leverage DUSt3R~\cite{wang2024dust3r} to predict the focal length of the images. The camera is assumed to follow a simple pinhole camera model, with the principle point fixed at the center of the image. When using predicted intrinsic parameters instead of ground-truth focal length, the localization accuracy shows a significant decline. However, enabling focal length refinement during EPnP-RANSAC~\cite{epnp,pnpsolver} yields notable improvement on $\textbf{Test}_{T1-T7}$, while maintaining similar performance on other two test splits. This suggests that refining predicted focal length during pose estimation is more effective when the correspondences are of higher quality.


\begin{table*}[t]
\centering
\begin{tabular}{c|c|c|c|c|c|c|c|c}
    & GT & Refine & \multicolumn{2}{c|}{$\textbf{Test}_{T1-T7}$} & \multicolumn{2}{c|}{$\textbf{Test}_{T1-T7 hard}$} & \multicolumn{2}{c}{$\textbf{Test}_{T10}$} \\ \cline{4-9}
    & Focal & Focal & RRE($^\circ$) & RTE($m$) & RRE($^\circ$) & RTE($m$) & RRE($^\circ$) & RTE($m$) \\ \hline
\multirow{3}{*}{Ours} & $\times$  & $\times$   & 2.04 & 1.72 & 4.56 & 2.19 & 4.61 & 4.80 \\ 
    & $\times$  & \checkmark   & 0.95 & 0.80 & 3.74 & 2.36 & 3.88 & 5.06  \\ 
    & \checkmark  & $\times$   & \textbf{0.66} & \textbf{0.51} & \textbf{2.64} & \textbf{1.13} & \textbf{2.53} & \textbf{2.69} \\ 
\end{tabular}
\caption{Ablation study on localization with intrinsic parameters predicted by DUSt3R~\cite{wang2024dust3r}. We report the median RRE and median RTE across all three test splits of the \textit{Carla Intersection} dataset. ``GT Focal'' refers to using the ground-truth focal length during inference, and ``Refine Focal'' enables focal length optimization as part of the EPnP-RANSAC~\cite{epnp,pnpsolver} process.}
\label{tab:supp_intrinsic}
\end{table*}

\noindent
\textbf{Block number of Fusion transformer.} Table \ref{tab:supp_ablation_attlayer} shows the experimental results of using different numbers of feature fusion layers $N_C$ in Geometry-guided Feature Fusion (GFF) module. Our model achieves the best performance when utilizing a four-layer structure.

\begin{table}[t]
\centering
\begin{tabular}{c|c|c|c}
    & $N_c$ & RRE($^\circ$) & RTE($m$) \\ \hline
\multirow{3}{*}{Baseline}    & 2   & 0.96    & 0.55   \\ 
    & 6  & 0.73    & 0.59   \\ 
    & 8  & 0.88    & 0.58   \\ \hline
Ours   & 4  & \textbf{0.66}   & \textbf{0.51}  \\ 
\end{tabular}
\caption{Ablation study on the number of feature fusion layers $N_c$ in Geometry-guided Feature Fusion (GFF) module. We report median RRE and median RTE on test split $\textbf{Test}_{T1-T7}$ of \textit{Carla Intersection} dataset.}
\label{tab:supp_ablation_attlayer}
\vspace{-0.2cm}
\end{table}

\noindent
\textbf{Input point cloud size.} We conducted ablation studies to investigate the effect of input point cloud size on the representation learning process. The number of coarse point groups was fixed to $M=512$, as these groups were generated using Farthest Point Sampling (FPS), ensuring uniform sampling across the point cloud. As shown in Table~\ref{tab:supp_ablation_pointsize}, the localization accuracy decreases with lower point cloud densities, as overly sparse point cloud lose local critical structural details. On the other hand, higher-density point clouds place a heavy computational burden. To balance computational efficiency and accuracy, we selected an input size of 20,480 points.

\begin{table}[t]
\centering
\begin{tabular}{c|c|c|c}
Point Number & RRE($^\circ$) & RTE($m$) & FLOPs \\ \hline
5120  & 0.86  & 0.68  & 126.73G  \\ 
10240 & 0.81  & 0.62  & 146.38G  \\ 
20480 & 0.66  & \textbf{0.51}  & 185.73G  \\ 
40960 & \textbf{0.59}  & 0.52  & 264.35G  \\ 
\end{tabular}
\caption{Ablation study on the input point cloud size. We report median RRE and median RTE on test split $\textbf{Test}_{T1-T7}$ of \textit{Carla Intersection} dataset. The FLOPs is calculated during the inference process.}
\label{tab:supp_ablation_pointsize}
\vspace{-0.2cm}
\end{table}

\begin{table*}[h]
\centering
\begin{tabular}{l|c|c|c|c}
    & \textbf{Training}& $\textbf{Test}_{T1-T7}$ & $\textbf{Test}_{T1-T7hard}$ & $\textbf{Test}_{T10}$ \\ \hline
worlds    & $Town01$-$07$  & $Town01$-$07$   & $Town01$-$07$   & $Town10$     \\ 
\# intersections & 67  & 7   & 7   & 1     \\ 
\# images per scene    & 768  & 288   & 288   & 288     \\ 
height (m)    & 6 / 7 / 8  & 6.5 / 7.5   & 6.5 / 7.5   & 6.5 / 7.5    \\ 
pitch ($^\circ$)    & 15 / 30  & 15 / 30   & 20 / 25   & 15 / 30    \\ \hline
seen intersection   & $-$  & $\times$    & $\times$   & $\times$   \\ 
seen world   & $-$  & \checkmark    & \checkmark   & $\times$   \\ 
\end{tabular}
\caption{Image data collection details of the proposed \textit{Carla Intersection} dataset. ``\# intersections'' means the number of intersection scenes in each split dataset and ``\# images per scene'' means the number of images in each intersection scene. ``Seen intersection'' and ``seen world'' represent whether the testing intersections are seen and whether the testing intersections are from the seen world during the training process, respectively.}
\label{tab:dataset_detail}
\vspace{-0.2cm}
\end{table*}

\section{Carla Intersection Dataset}
\label{sec: supp_carla_dataset}
Our proposed \textit{Carla Intersection} dataset consists of 75 intersections across 8 worlds ($Town01$ to $Town07$ and $Town10$) within the CARLA~\cite{Carla} simulation environment, encompassing both urban and rural landscapes. $Town01$ to $Town07$ include multiple intersections for training and testing, while $Town10$ contains only one intersection for testing. Specifically, we utilize the first Intersection scenario from each world 
(e.g. $Town01$ $Int1$, $Town02$ $Int1$, \dots, $Town07$ $Int1$, $Town10$ $Int1$) 
for testing, with all remaining intersections reserved for training.

\begin{figure}[t]
    \centering
    \includegraphics[width=1.0\linewidth]{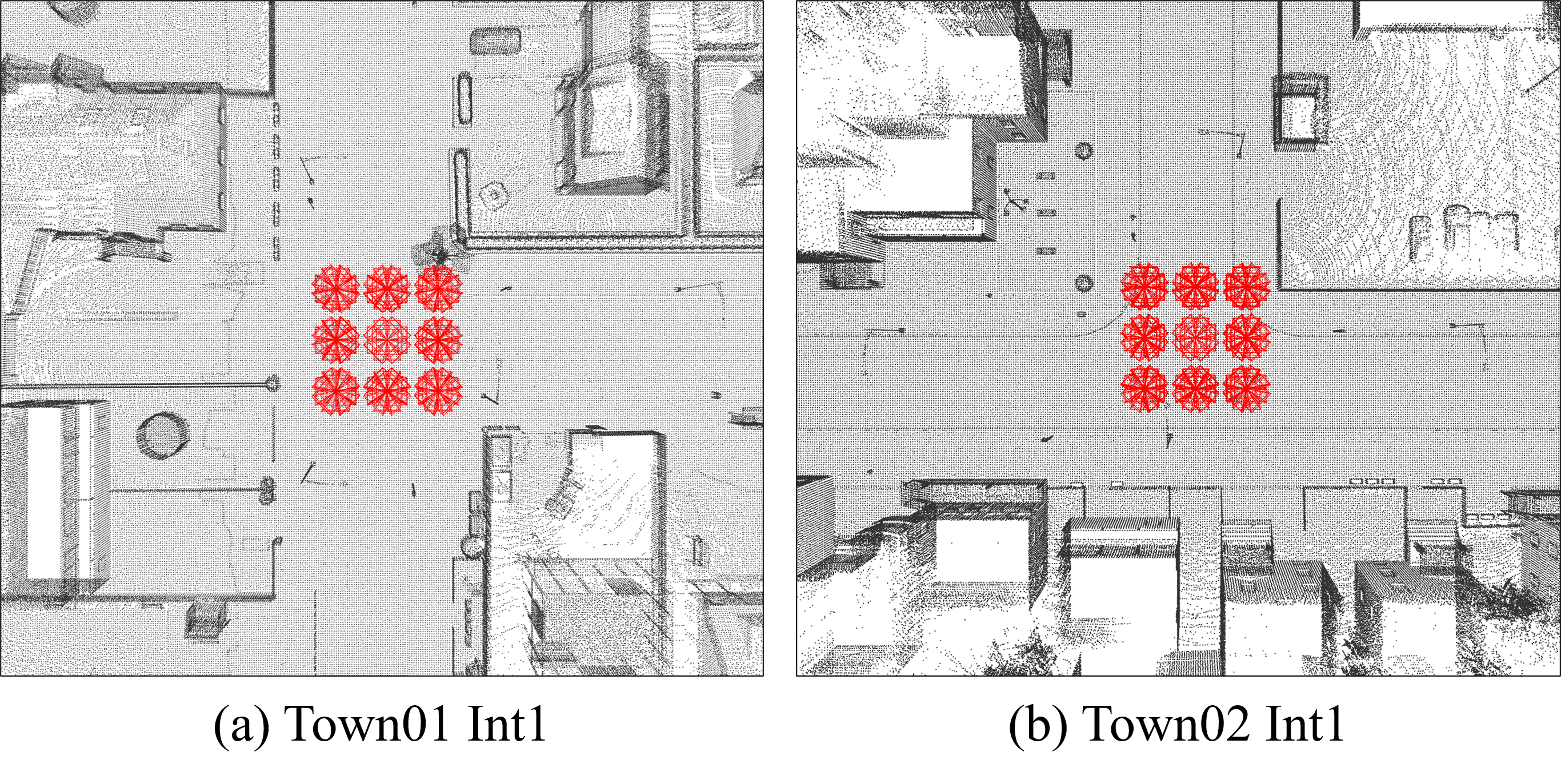}
    \caption{Sampled testing image poses of (a) Town01 Intersection1 and (b) Town02 Intersection1.}
    \label{fig:sampled_pose}
    \vspace{-0.2cm}
\end{figure}

\noindent
\textbf{Images.} 
For each intersection, we captured 768 training images and 288 testing images with known ground-truth 6-DoF pose at a resolution of 1920x1080 pixel and a horizontal field of view (FOV) of $90^\circ$, equals to a focal length of 960. 
To generate these images, we sampled camera positions in a grid-like pattern with different heights at the center of each intersection. For each position, we captured images at 8 yaw angles (spaced at 45$^\circ$ intervals) and 2 pitch angles. Figure \ref{fig:sampled_pose} shows the sampled poses for example intersections. 

Table \ref{tab:dataset_detail} summarizes the image data collection details for our \textit{Carla Intersection} dataset. All training images were captured with downward pitch angles of $15^\circ$ and $30^\circ$ at heights of \SI{6}{m}, \SI{7}{m}, and \SI{8}{m}. Testing images in the test splits $\textbf{Test}_{T1-T7}$ and $\textbf{Test}_{T10}$ share the same pitch angles as the training images, but were captured at heights of \SI{6.5}{m} and \SI{7.5}{m}. Additionally, for the test split $\textbf{Test}_{T1-T7hard}$, we captured 288 additional testing images for each intersection using the same positions as in $\textbf{Test}_{T1-T7}$, but with different pitch angles of $20^\circ$ and $25^\circ$, at heights of \SI{6.5}{m} and \SI{7.5}{m}. These data capture settings closely reflect the real-world traffic surveillance camera installations following HIKVISION~\cite{hikvision}, ensuring typical positioning to provide optimal traffic views under varied monitoring conditions. 
The differences between three distinct test splits also allow us to evaluate the model's generalization ability across unseen intersections and unseen world styles. Note that all testing intersections were not seen during the training.

\begin{figure*}[t]
    \centering
    \includegraphics[width=0.9\linewidth]{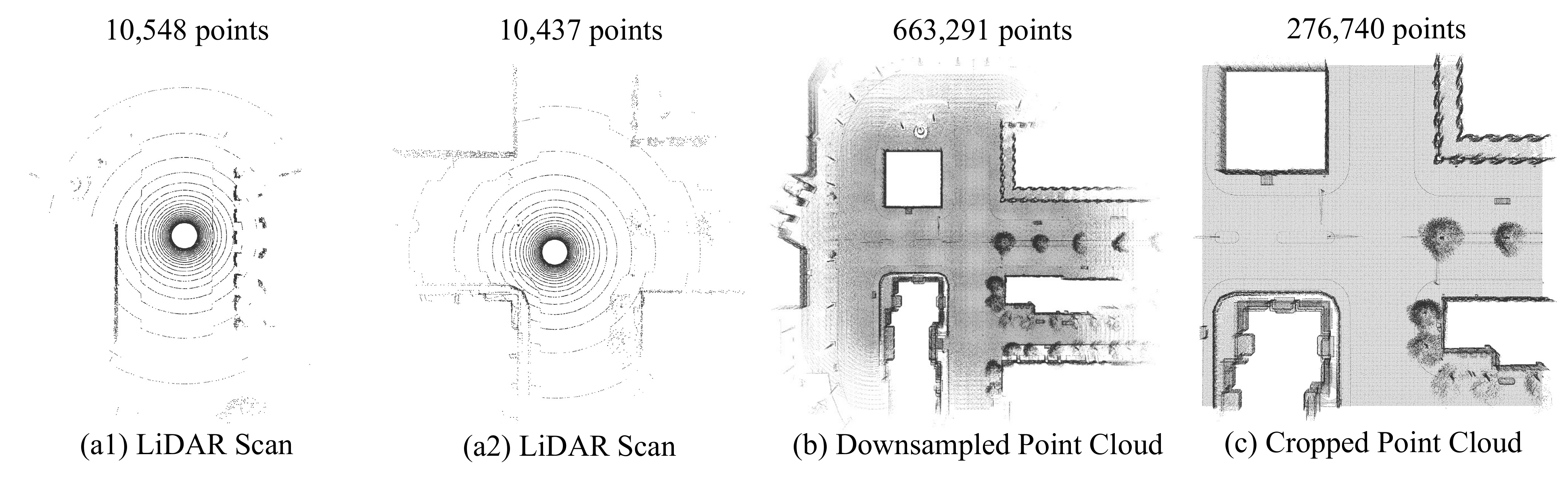}
    \caption{Point cloud capturing example from $Town10$ $Int1$. (a1) and (a2) depict the LiDAR scan from a single frame. (b) shows the aggregated and downsampled point cloud. (c) presents the final cropped point cloud with dimensions of \SI{100}{m}$\times$\SI{100}{m}$\times$\SI{100}{m}.}
    \label{fig:lidar_scan}
    \vspace{-0.5cm}
\end{figure*}

\noindent
\textbf{Point Clouds.}
To capture the point cloud of each intersection, we utilize a simulated LiDAR sensor in the CARLA~\cite{Carla} environment, which emulates a rotating LiDAR using ray-casting. The LiDAR operates at a rotation frequency of 10 frames per second (FPS), with a vertical field of view (FOV) ranging from 10$^\circ$ (upper) to -30$^\circ$ (lower). The sensor generates 224,000 points per second across all lasers. Other parameters of the simulated LiDAR follow the default configuration in CARLA Simulator. 
As shown in Figure \ref{fig:lidar_scan}, the LiDAR scans were captured in an on-board manner. Then, we accumulated all scans into a single point cloud and downsampled it with a resolution of \SI{0.2}{m}. Finally, the point cloud for each intersection was cropped to a region measuring \SI{100}{m}$\times$\SI{100}{m}$\times$\SI{50}{m}, focusing on the area of interest for our study.

\begin{figure*}[t]
    \centering
    \includegraphics[width=1.0\linewidth]{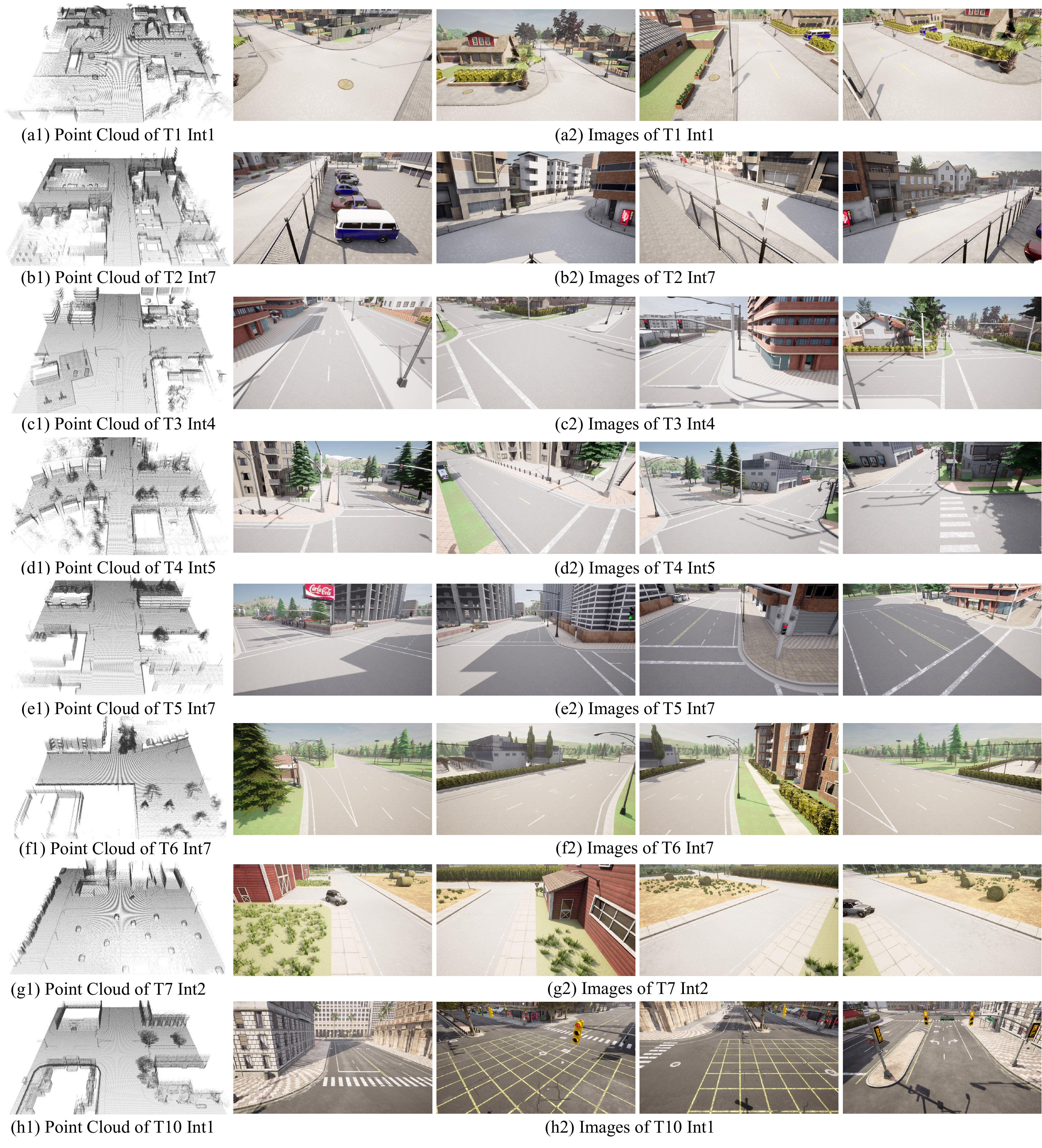}
    \caption{Example point clouds and images data of our \textit{Carla Intersection} dataset. T1 means $Town01$ and Int1 means $Intersection 1$. Since all instances of the $Intersection1$ scenario across different worlds are included in the test set, we focus on showcasing their testing images (e.g. T1 Int1 and T10 Int1). For other intersections, we present the training images instead.}
    \label{fig:carla_dataset}
    \vspace{1.5cm}
\end{figure*}

During the data capturing process, we disabled dynamic weather variations and set the weather condition in CARLA simulation environment to the default weather parameters of world $Town10$. Some examples of our \textit{Carla Intersection} dataset are shown in Figure \ref{fig:carla_dataset}. Our data collection codes and datasets will be publicly available upon acceptance. 

\section{Geometry-guided Feature Fusion}
\label{sec: supp_gff}
Our Geometry-guided Feature Fusion (GFF) module comprises of $N_{c}$ transformer-based fusion blocks, each consisting of a self-attention layer followed by a cross-attention layer. 

Given the image feature $\mathbf{F}_I$ and point cloud feature $\mathbf{F}_P$, both enriched with positional embeddings, the self-attention layer enhances features within each modality individually using standard multi-head scalar dot-product attention:
\begin{equation}
    \dot{\mathbf{F}}=\mathbf{Q}+\operatorname{MHA}(\mathbf{Q}, \mathbf{K}, \mathbf{V}),
    \label{equation: MHA}
\end{equation}
where $\mathbf{Q}=\mathbf{K}=\mathbf{V}=\mathbf{F} \in \mathbb{R}^{N_t \times C}$ denotes the \textit{Query, Key} and \textit{Value} matrices, and $\mathbf{F}$ represents either $\mathbf{F}_I$ or $\mathbf{F}_P$ depending on the modality. 
Within the MHA layer, the attention operation is conducted by projecting $\mathbf{Q}, \mathbf{K}$ and $\mathbf{V}$ using $h$ heads:
\begin{equation}
\begin{aligned}
    & \operatorname{MHA}(\mathbf{Q}, \mathbf{K}, \mathbf{V}) = [head_{1},\dots,head_{h}]\mathbf{W}^{O} \\
    & head_{i} = \operatorname{Attention}(\mathbf{Q}\mathbf{W}^{Q}_{i}, \mathbf{K}\mathbf{W}^{K}_{i}, \mathbf{V}\mathbf{W}^{V}_{i})
\end{aligned}
\end{equation}
where $\textbf{W}_{i}^{Q,K,V,O}$ denote the learnable parameters of linear projection matrices and the $\operatorname{Attention}$ operation is defined as:
\begin{equation}
    \operatorname{Attention}(q, k, v)=\operatorname{Softmax}(\frac{q \cdot k^{\top}}{\sqrt{d_k}})v,
\end{equation}
where $d_k$ is the dimension of latent feature. 

The cross-attention layer fuses image and point cloud features by applying the attention mechanism across modalities, following the same formulation as Equation~\ref{equation: MHA}. However, the \textit{Query, Key} and \textit{Value} matrices differs based on the direction of attention. Specifically, for \textbf{I2P} (Image-to-Point Cloud) attention, we use $\mathbf{Q}=\mathbf{F}_I$ and $\mathbf{K}=\mathbf{V}=\mathbf{F}_P$, while for \textbf{P2I} (Point Cloud-to-Image) attention, we set $\mathbf{Q}=\mathbf{F}_P$ and $\mathbf{K}=\mathbf{V}=\mathbf{F}_I$.

Layer Normalization is applied to ensure stable training. For our GFF module, we set $N_{c}=4$ and $h=4$. Both the input channel $C$ and the latent dimension $d_k$ are set to 256.

\begin{figure*}[t]
    \centering
    \includegraphics[width=1.0\linewidth]{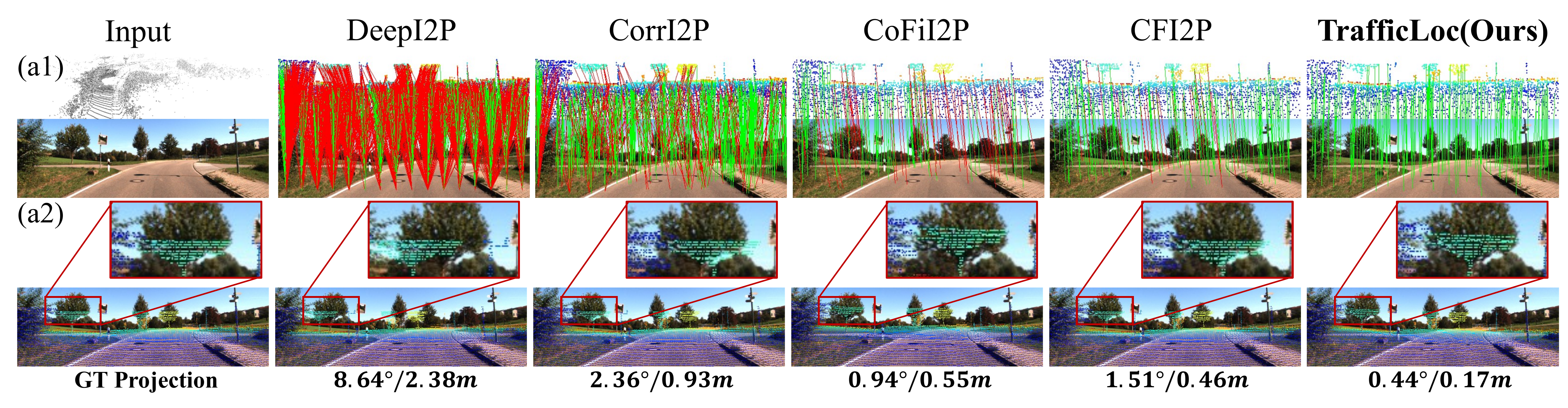}
    \caption{Qualitative results of our TrafficLoc and other baseline methods on the KITTI Odometry dataset~\cite{kitti}. (a1) shows predicted correspondences and (a2) visualizes the point cloud projected onto the image plane. The first column provides the input point cloud, the input image and the ground-truth projection for reference.}
    \label{fig:supp_result_kitti}
\end{figure*}

\section{Implementation Details}
\label{sec: supp_implement}
In \textit{Carla Intersection} dataset, each intersection point cloud represents a region of \SI{100}{m}$\times$\SI{100}{m}$\times$\SI{50}{m} and contains over 200,000 points. Following~\cite{tang2023neumap}, as a preprocessing step, we first divide each intersection point cloud into several \SI{50}{m}$\times$\SI{50}{m}$\times$\SI{50}{m} voxels with a stride of $25m$. For each voxel $V_i$, we assign an associated set of images $\{I_i\}$ based on the overlap ratio between the image frustum and the voxel. Specifically, a voxel $V_i$ is associated with an image $I_i$ if more than 30\% projected points lie within the image plane. During each training epoch, we uniformly sample $B$ images for each voxel from its associated image set, resulting in $B\cdot N_v$ training image-point cloud pairs, where $N_v$ denotes the total number of voxels. 

The input images are resized to 288 $\times$ 512, and the input point cloud size is $N=20480$ points. We utilize a pre-trained Vision Transformer from \textit{DUSt3R\_ViT Large}~\cite{wang2024dust3r} to extract the image feature. For coarse matching, we use a resolution of 1/16 of the input resolution for image ($s=16$) and set the number of point group $M=512$, with a coarse feature channel size of $C=256$. For fine matching, we adopt a resolution of $(H/2\times W/2\times C^\prime)$ for fine image feature and $(N\times C^\prime)$ for fine point feature, where $H$, $W$ and $N$ equal to input dimensions and the fine feature channel size is set to $C^\prime=64$. As part of data augmentation, we apply random center cropping to the input images before resizing operation to simulate images captured by different focal lengths. The input point cloud is first normalized into a unit cube,
followed by random rotations around the z-axis (up to 360$^\circ$) and random shifts along the xy-plane (up to \SI{0.1}{m}).

The whole network is trained for 25 epochs with a batch size of 8 using the Adam optimizer~\cite{adam}. The initial learning rate is set to 0.0005 and is multiplied by 0.5 after every 5 epochs. For the joint loss function, we set $\lambda_1=\lambda_2=\lambda_3=\lambda_4=1$. The safe radius $r$, positive margin $m_p$, negative margin $m_p$ and scale factor $\gamma$ in loss function are set to 1, 0.2, 1.8 and 10, respectively. For the Geometry-guided Attention Loss (GAL), the angular thresholds $\theta_{low}$ and $\theta_{up}$ are set to 10$^\circ$ and 20$^\circ$, while the distance thresholds $d_{low}$ and $d_{up}$ are set to 3m and 5m, respectively. The training is conducted on a single NVIDIA RTX 6000 GPU and takes approximately 40 hours.

During inference, we utilize the super-point filter to select reliable in-frustum point groups from the fused coarse point features $\mathbf{F}^{coarse}_P$, using a confidence threshold of 0.9. In the \textbf{coarse matching stage}, we compute the coarse similarity map between each point group and the image. Following~\cite{zhang2024telling}, a \textbf{window soft-argmax} operation is employed on similarity map to estimate the corresponding coarse pixel position. This involves first identifying the target center with an argmax operation, followed by a soft-argmax within a predefined window (window size set to 5). 
In the \textbf{fine matching stage}, with the predicted coarse pixel position, we first extract a fine local patch feature of size $w\times w$ from the fine image feature and select the fine point feature of each point group center, and then compute the fine similarity map between each point group center and the extracted local patch. 
Since the extracted local fine image patch has a relative small size ($w=8$), a soft-argmax operation is applied over the \textbf{entire} fine similarity map to determine the final corresponding 2D pixel for each 3D point group center.
Finally, we estimate the camera pose using EPnP-RANSAC~\cite{epnp, pnpsolver} based on the predicted 2D-3D correspondences. For cases where one single image is associated with multiple point clouds, an additional EPnP-RANSAC step is performed using all inliers from each image-point cloud pair to compute the final camera pose. 

For experiments on the KITTI Odometry~\cite{kitti} and Nuscenes~\cite{NuScenes} datasets, we ensure a fair comparison by adopting the same procedures as in previous works~\cite{deepi2p, ren2022corri2p, kang2023cofii2p} to generate image-point cloud pairs.

In the KITTI Odometry dataset~\cite{kitti}, there are 11 sequences with ground-truth camera calibration parameters. Sequences 0-8 are used for training, while sequences 9-10 are reserved for testing. Each image-point cloud pair was selected from the same data frame, meaning the data was captured simultaneously using a 2D camera and a 3D LiDAR with fixed relative positions. During training, the image resolution was set to 160$\times$512 pixels, and the number of points was fixed at 20480. The model was trained with a batch size of 8 until convergence. The initial learning rate is set to 0.001 and is multiplied by 0.5 after every 5 epochs.

For the NuScenes dataset~\cite{NuScenes}, we utilized the official SDK to extract image-point cloud pairs, with the point clouds being accumulated from the nearby frames. The dataset includes 1000 scenes, of which 850 scenes were used for training and 150 for testing, following the official data split. The image resolution was set to 160$\times$320 pixels, and the number of points was fixed at 20480.

\section{More Visualization Results}
\label{sec: supp_visual}
\begin{figure*}[t]
    \centering
    \includegraphics[width=1.0\linewidth]{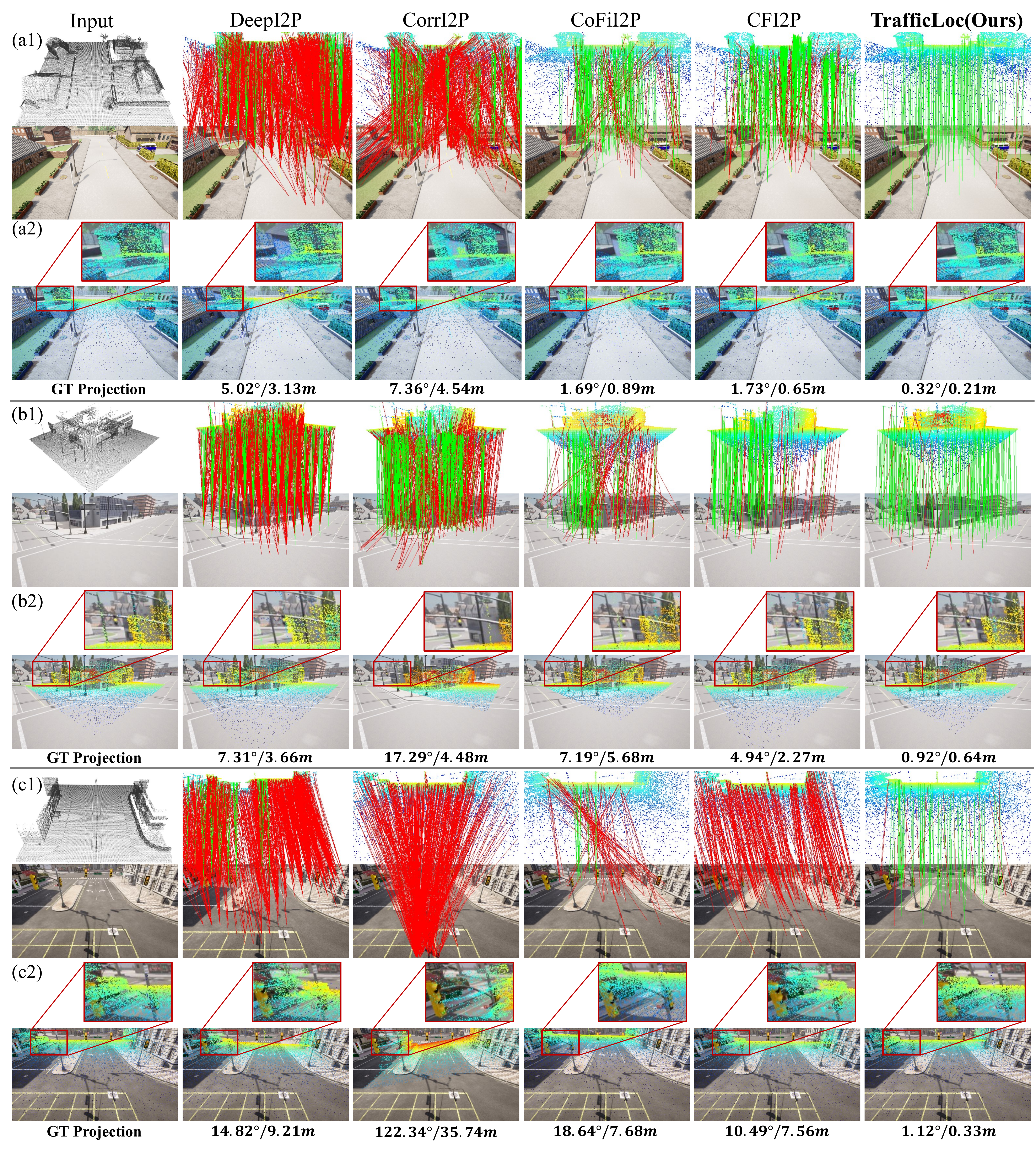}
    \caption{Qualitative results of our TrafficLoc and other baseline methods on the \textit{Carla Intersection} dataset. The point cloud is projected onto a 2D view and displayed above the image, with point colors indicating distance. The proposed TrafficLoc achieves superior performance, with more correct (\textcolor{green}{green}) and fewer incorrect (\textcolor{red}{red}) point-to-pixel pairs. (a1) shows predicted correspondences on $\textbf{Test}_{T1-T7}$ and (a2) visualizes the point cloud projected onto the image plane. Similarly, (b1) and (b2) show results on $\textbf{Test}_{T1-T7 hard}$, (c1) and (c2) show results on $\textbf{Test}_{T10}$. The first column provides the input point cloud, the input image and the ground-truth projection for reference.}
    \label{fig:supp_result_carla}
\end{figure*}

In this section, we present additional examples of localization results. 
Figure~\ref{fig:supp_result_kitti} and Figure~\ref{fig:supp_result_carla} compare the localization performance of TrafficLoc with other baseline methods on the KITTI Odometry dataset~\cite{kitti} and all three test splits of the \textit{Carla Intersection} dataset, respectively. Our TrafficLoc predicts a higher number of correct point-to-pixel correspondences, and the point cloud projected with the predicted pose exhibits greater overlap with the image, demonstrating superior performance.


\end{document}